\documentclass[10pt,twocolumn,letterpaper]{article}

\usepackage{iccv}     
\toggletrue{iccvpagenumbers}

\usepackage[accsupp]{axessibility}  %
\usepackage{booktabs}
\usepackage{multirow}
\usepackage[english]{babel}
\usepackage{microtype}
\usepackage{dcolumn} %
\newcolumntype{d}[1]{D{.}{.}{#1}}
\usepackage{pifont}
\newcommand{\cmark}{\ding{51}}%
\usepackage[absolute,overlay]{textpos}

\definecolor{iccvblue}{rgb}{0.21,0.49,0.74}
\definecolor{emerald}{rgb}{0.31, 0.78, 0.47}
\usepackage[pagebackref,breaklinks,colorlinks,allcolors=iccvblue]{hyperref}

\def\confName{ICCV}
\def\confYear{2025}

\title{How Would It Sound?\\ Material-Controlled Multimodal Acoustic Profile Generation for Indoor Scenes}

\author{
Mahnoor Fatima Saad \quad Ziad Al-Halah\\
University of Utah\\
{\tt\small \{mahnoor.saad, ziad.al-halah\}@utah.edu}
}

\begin{document}

\twocolumn[
    \vspace{-1.5em}
  \begin{center}
    {\small In Proceedings of IEEE/CVF International Conference on Computer Vision (ICCV), 2025.}
    \vspace{-2.15em}
  \end{center}
  \maketitle
]

\thispagestyle{plain}
\pagestyle{plain}

\begin{abstract}

How would the sound in a studio change with a carpeted floor and acoustic tiles on the walls? We introduce the task of \textit{material-controlled acoustic profile generation}, where, given an indoor scene with specific audio-visual characteristics, the goal is to generate a target acoustic profile based on a user-defined material configuration at inference time. We address this task with a novel encoder-decoder approach that encodes the scene's key properties from an audio-visual observation and generates the target Room Impulse Response (RIR) conditioned on the material specifications provided by the user. Our model enables the generation of diverse RIRs based on various material configurations defined dynamically at inference time. 
To support this task, we create a new benchmark, the \textit{Acoustic Wonderland Dataset}, designed for developing and evaluating material-aware RIR prediction methods under diverse and challenging settings. Our results demonstrate that the proposed model effectively encodes material information and generates high-fidelity RIRs, outperforming several baselines and state-of-the-art methods. Project: \url{https://mahnoor-fatima-saad.github.io/m-capa.html}
\end{abstract}

\section{Introduction}
\label{sec:intro}

\begin{figure}
    \centering
    \includegraphics[width=1\linewidth]{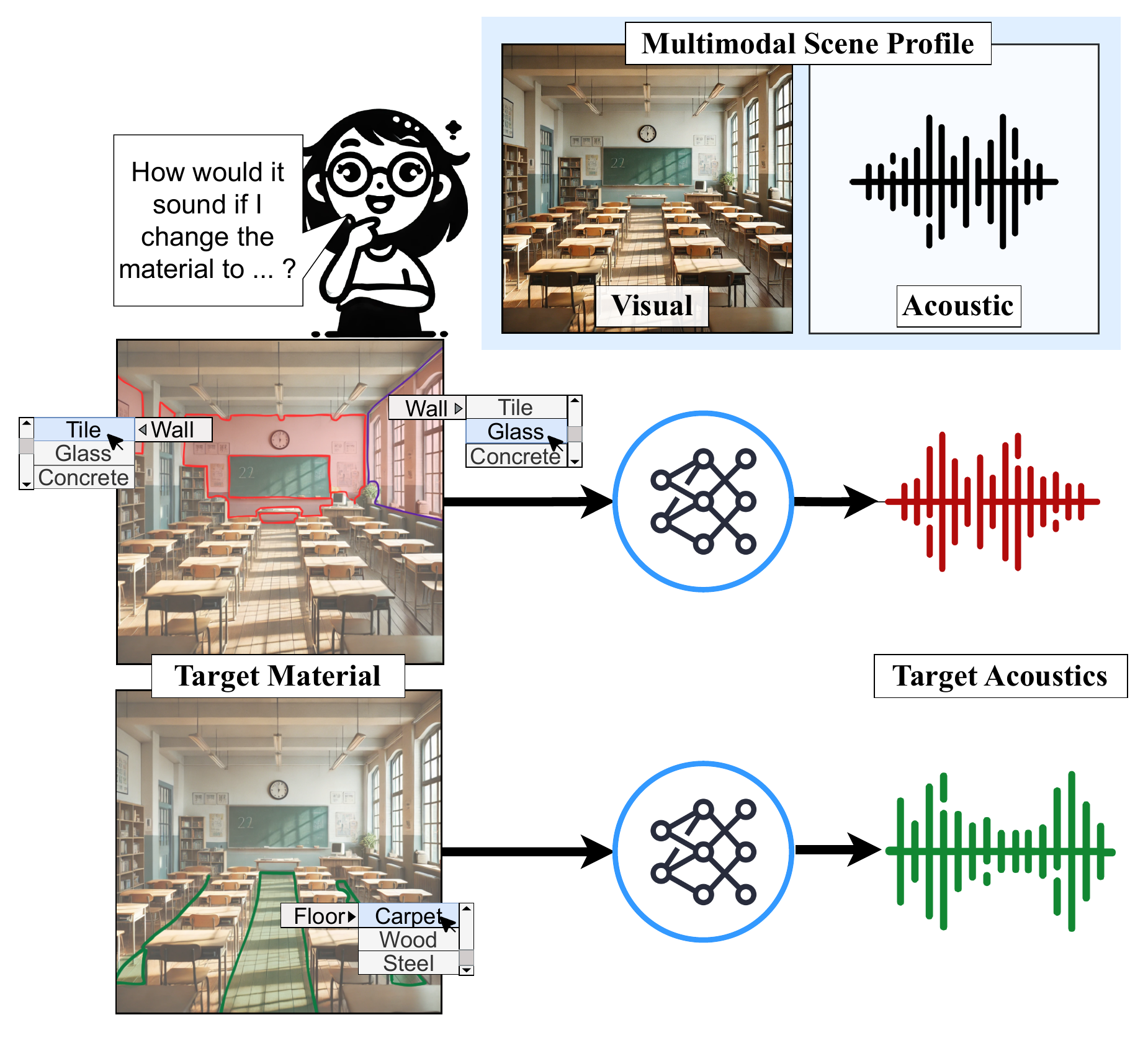} %
    \vspace{-0.5cm}
    \caption{
        Our material-controlled acoustic profile generation task. Given a scene with specific visual and acoustic properties (top), the objective is to create new acoustic profiles for the same scene by allowing users to dynamically assign different materials to objects. Our approach enables users to experience how various material configurations impact the sound quality of the scene on the fly. For instance, a user can explore how a classroom would sound with \emph{acoustic tiles} and big \emph{glass} windows on the walls (middle) or a \emph{carpeted} floor (bottom).
    }
    \vspace{-0.2in}
    \label{fig:intro}
\end{figure}

Sound, along with vision, plays a fundamental role in shaping our perception of the environment. From conveying essential spatial information to enhancing emotional and social experiences, sound improves our ability to interpret and interact with the world around us. Realistic sound modeling is therefore essential in applications that aim to create immersive, lifelike experiences, such as AR/VR and gaming, where a mismatch between the visual and acoustic stimuli may lead to the \textit{room divergence effect}~\cite{werner2021creation} and the collapse of the plausibility of the whole experience. 

Achieving accurate and immersive audio experiences requires precise modeling of how sound propagates in a given space. As sound travels through a room, it interacts with objects, surfaces, and materials through processes like reflection, absorption, and transmission. These interactions impart unique reverberations to the sound wave that are specific to each environment, creating a distinct acoustic signature. For example, the experience of listening to a symphony in a living room differs dramatically from listening in a theater hall. The Room Impulse Response (RIR) function~\cite{vigran2014building} captures these unique characteristics by modeling how sound travels between two points within a space. By convolving an audio signal with an RIR, we can reproduce the acoustic characteristics of the room, producing a sound that closely resembles what a listener would experience in that environment.

Given the importance of accurate RIR modeling, there is considerable interest in developing high-fidelity RIR prediction methods~\cite{IR-GAN, Fast-RIR}. When the geometry, object distribution, and materials of a room are known and represented in a 3D mesh, simulation methods~\cite{FDTD, 10.1145/2980179.2982431, 10.1121/1.398336,7849179}, such as ray tracing, can be used to measure the RIR. However, creating such detailed 3D meshes requires expensive measurements and time-consuming labeling, limiting the scalability of these approaches to diverse scenarios. Recent work has shifted toward predicting RIRs using sparse and inexpensive data sources, such as room dimensions~\cite{inras, naf, Fast-RIR},  images~\cite{NACF, AV-NERF, AV-RIR, naf}, or recorded audio~\cite{DBLP:journals/corr/abs-1911-06245, Fast-RIR}. While these approaches show promise in generalizing across scenes, they often simplify room representations, typically modeling them as simple boxes~\cite{inras} or just via an RGB image   {~\cite{image2reverb}}. Consequently, the materials within a scene are frequently overlooked, and the model must infer material properties implicitly from RGB images.

However, material properties have a significant impact on the RIR of a space~\cite{AV-RIR}. Even in a room with identical geometry and object placement, the perceived sound can vary substantially depending on whether, for example, walls are made of wood, concrete, or soundproof materials. Different materials interact with sound in distinctive ways, modifying its behavior by dampening, amplifying, or introducing specific reverberations across various frequencies. A few recent methods have incorporated explicit material representations in RIR prediction~\cite{AV-RIR, Listen2Scene}, with \cite{AV-RIR} showing that including material properties in model inputs leads to more accurate RIR predictions. However, none of these existing methods provide users with the flexibility to adjust the material configuration of a scene at inference time to generate an RIR that reflects such changes.

To address this challenge, we introduce the new task of \textit{material-controlled acoustic profile generation} (see Fig.~\ref{fig:intro}). In this task, the goal is to generate an RIR that reflects a hypothetical scene configuration, where an initial audio-visual observation of the scene provides the scene's original characteristics, and a user-defined material configuration specifies the new material assignments for objects and surfaces.

The ability to control material configurations in RIR prediction has valuable practical applications across domains such as VR/AR, creative design, and architectural engineering. This capability enables users to make informed decisions based on simulated acoustics for different material setups. For example, an instructor could evaluate how a classroom would sound if its walls were covered in wood; a music enthusiast could experience the acoustic effects of their studio with a carpeted floor or large glass windows; and interior designers could assess the impact of various materials on furniture and objects to enhance a room's acoustics. All of this can be done without physically altering the space or purchasing expensive materials.

To tackle this task, we present a novel approach that encodes the scene's initial properties from audio-visual data and enables the user to define an arbitrary material mask, allowing them to assign specific materials to selected objects in the scene. Our model processes the new material configuration alongside the original scene representation to generate a target RIR, using an encoder-decoder architecture designed to capture how the new material configuration will influence the RIR.

Furthermore, to support research on this task, we introduce a new dataset, \textit{Acoustic Wonderland Dataset}, designed to model the impact of material properties on RIR predictions explicitly, leveraging state-of-the-art audio-visual simulators~\cite{habitat, chen22soundspaces2}. Our benchmark evaluates model performance across different generalization scenarios, including seen and unseen material configurations and room geometries. Our evaluation on this challenging benchmark demonstrates the effectiveness of our approach, outperforming various baselines and existing methods that incorporate material information either implicitly or explicitly. Furthermore, we conduct a user study that demonstrates our model's ability to generalize well to real-world scenarios.

\section{Related Works}
\label{sec:related}

Estimating room impulse responses (RIRs) in a 3D scene has numerous applications, like augmented and virtual reality (AR/VR)~\cite{liu2021soundsynthesispropagationrendering,kim2019immersive}, audio-visual navigation~\cite{chen20soundspaces,Chen_2021_CVPR,9197008}, speech enhancement~\cite{AV-RIR, Adverb,ratnarajah2023improvedroomimpulseresponse}, and audio-visual localization~\cite{Wu_Wu_Ju_Wang_2021,mo2023avsamsegmentmodelmeets,Tian_2018_ECCV}. In this section, we review key directions in the literature on RIR estimation relevant to this work.

\vspace{-0.5cm}
\paragraph{Physics- and Geometry-based RIR Modeling}
Traditional methods estimate RIRs by using physics-based equations to model acoustic wave propagation~\cite{FDTD, 10.1121/1.2164987, 10.1121/1.3021297, 5165582, MEHRA201283}, or by applying geometry-based methods such as ray tracing~\cite{10.1145/2980179.2982431, 10.1121/1.398336, 10.1121/1.382599}. However, these methods often require extensive manual measurements~\cite{5753892} or make simplified assumptions about the environment~\cite{9414399} (e.g., approximating it as a rectangular box).
Machine learning methods have shown promising results for RIR estimation. Leveraging advanced audio-visual simulators~\cite{schissler2016interactive,chen20soundspaces,chen22soundspaces2}, these methods train deep neural networks to estimate RIRs for any given source and receiver location within complex environments~\cite{naf}. However, such methods typically require access to the full 3D mesh of the scene~\cite{inras, naf,ratnarajah2021ts} or user-provided scene geometry~\cite{NACF}, making them computationally expensive~\cite{schissler2016interactive} and limiting their generalizability to novel scenes. In this work, we propose a method for RIR modeling based on multimodal observations from a single location, alleviating the need for scene meshes or explicit geometric properties and capable of generalization to novel environments.

\vspace{-.5cm}
\paragraph{RIRs from Audio-Visual Observations}
Recent approaches have aimed to bypass full 3D scene modeling by using limited or single multimodal observations to estimate RIRs. Early methods used scene images to estimate only the late reverberant characteristics of an RIR~\cite{kon2018deep,kon2020auditory} or to infer room geometry from panoramic images, subsequently synthesizing RIRs based on these estimates~\cite{remaggi2019reproducing}. Image2Reverb~\cite{image2reverb} improved on this by generating full RIRs directly from RGB and depth inputs, while other approaches~\cite{chen2022visual,chen2023novel,somayazulu2024self} used audio-visual observations for implicit RIR modeling, tailoring acoustics to a specific scene~\cite{chen2022visual,somayazulu2024self} or even a novel viewpoint~\cite{chen2023novel}. 
Further methods have sought to predict RIRs for arbitrary locations within a scene using a few images and acoustic responses~\cite{majumder2022fsrir,NACF}, demonstrating improved performance compared to purely geometry-based methods~\cite{Fast-RIR,naf,inras}. Despite notable advancements in high-fidelity RIR generation, these methods do not explicitly model objects and surface materials, which limits the accuracy of the generated RIRs~\cite{AV-RIR}.

\vspace{-.5cm}
\paragraph{Explicit Material Modeling}
The materials present in a scene significantly influence its acoustic properties, as different materials have unique absorption, transmission, and reflection characteristics that directly affect the RIR. 
Some methods have incorporated explicit material modeling in RIR generation~\cite{AV-RIR,Listen2Scene,schissler2017acoustic,li2018scene}, achieving superior performance over material-agnostic approaches~\cite{AV-RIR}. However, they typically require dense observation sampling and 3D mesh reconstruction to estimate materials~\cite{schissler2016interactive,li2018scene,tang2022gwa,Listen2Scene}, they rely on predefined mappings between semantic categories and material types~\cite{AV-RIR,Listen2Scene} (e.g., all walls are brick, all chairs are wood), or retrieve material-related late reverberations from the training set~\cite{AV-RIR}. As a result, these methods struggle to generalize when RIRs must be estimated for new scenes with novel material configurations. 
In contrast, our method explicitly models materials in RIR generation and can adjust RIR predictions based on new material configurations during inference, using only a single audio-visual observation from the scene. To our knowledge, this is the first work to address material-controlled RIR generation with arbitrary material configurations at inference time.

\section{Material-Controlled RIR Generation}
\label{sec:approach}

We present the first work to address the task of Room Impulse Response (RIR) generation, conditioned on arbitrary material configurations at inference time. Next, we begin by formally defining this novel task (Sec.~\ref{sec:task}), followed by a description of the dataset collected for this purpose (Sec.~\ref{sec:dataset}). Finally, we introduce our new approach for high-fidelity, material-controlled RIR generation (Sec.~\ref{sec:model}).

\subsection{Task Definition}
\label{sec:task}

The goal of our novel task is to predict the changes in an RIR of a 3D scene given a new material configuration provided at inference time. Specifically, while the scene's geometry, surfaces, and objects remain unchanged, the user can modify the material properties of these elements at inference (e.g., assign \emph{wood} to walls), and our goal is to anticipate the changed RIR accordingly.

Formally, let $\mathcal{S}$ denote a 3D scene. From a random location $l$ with coordinate $(x, y)$ and orientation $\theta$ in $\mathcal{S}$, we sample a multimodal observation $O = (V, A)$, where $V$ is an egocentric visual view of $\mathcal{S}$ from $l$, represented as an RGB image, %
and $A$ represents the RIR of a binaural echo response from $l$. Given a target material configuration $\mathcal{M}_T$ for $\mathcal{S}$, our goal is to predict the target RIR, ${A}_T$, consistent with the specified $\mathcal{M}_T$. Formally, we aim to learn a mapping ${A}_T = f((V, A); \mathcal{M}_T)$, where the user can define multiple, distinct $\mathcal{M}_T$ configurations for a given observation $O$ and generate their corresponding $A_T$ each.

In this work, we represent $\mathcal{M}_T$ using a segmentation mask, derived from a semantic segmentation mask $G$ inferred from $V$. This representation provides a flexible and intuitive interface for defining $\mathcal{M}_T$, allowing the user to simply click on an object or surface $c_i$ in $G$ and assign it a material class $m_j$, as demonstrated in Fig.~\ref{fig:intro}. This method eliminates the need for pixel-wise material assignments. Unselected areas or objects in $G$ are assigned an \emph{unchanged} material class, $m_u$, in $\mathcal{M}_T$. While we adopt this representation of $\mathcal{M}_T$ in this work, alternative approaches, such as language-based queries (e.g., "assign \emph{ceramic} to tables"), could also be explored and are left for future work.

\subsection{Acoustic Wonderland Dataset}
\label{sec:dataset}

To our knowledge, there are no publicly available datasets compatible with our task. Therefore, we introduce a novel dataset, named the Acoustic Wonderland dataset, which we discuss next (see the Supp for more details). %

\vspace{-0.4cm}
\paragraph{Platform and Scenes}
We use the SoundSpaces 2.0 (SSv2)~\cite{chen22soundspaces2} audio-visual 3D simulator to collect our dataset. SSv2, built on the AI-Habitat platform~\cite{habitat}, is a state-of-the-art simulator that offers fast and realistic audio-visual rendering, shown to transfer effectively to real-world settings~\cite{chen2024sim2real}. Additionally, we use $84$ Matterport3D (M3D) scenes~\cite{Matterport3D}, comprising 3D meshes derived from scans of real-world homes and indoor spaces. This enables us to evaluate our approach across numerous environments and diverse material configurations, facilitating comparisons with multiple baselines under consistent, reproducible conditions while simulatenously using realistic audio-visual renderings that closely resemble real-world scenes. %

\vspace{-0.4cm}
\paragraph{Material Profiles}
SSv2 applies predefined material-object mappings when rendering audio-visual observations, with each material characterized by its absorption, transmission, and reflection coefficients. SSv2 includes $30$ material definitions and there are over $40$ semantic categories in M3D. To balance storage efficiency with comprehensive material representation in our dataset, we select a representative set of $12$ material classes, $M=\{m_i\}$ (e.g., \emph{wood}, \emph{concrete}, \emph{steel}, \emph{soundproof}), and identify a subset of semantic categories $C=\{c_j\}$ representing prominent objects and surfaces (e.g., ceiling, floor, tables). We then generate a set of random mappings between $C$ and $M$ (material profiles) such that each $c_j$ is randomly assigned to a material $m_i$. For small or infrequent objects (e.g., ball, shoes), we retain the default SSv2 material mappings. In total, we create $2,673$ material profiles, $\mathcal{P} = \{\mathcal{P}_k\}$.

\vspace{-0.4cm}
\paragraph{Observation Sampling}
For each scene $\mathcal{S}_i$ in M3D, we sample $N$ random locations $l_n = (x_n, y_n, \theta_n)$ from spatial coordinates $(x_n, y_n)$ and orientation $\theta_n$. At each location $l_n$, we capture the RGB view, $V_n$, and the corresponding semantic segmentation mask $G_n$. Furthermore, at each $l_n$, we initialize SSv2 $J$ times with random material profiles $\mathcal{P}_j \in \mathcal{P}$, generate the corresponding material segmentation mask $\mathcal{M}_{n,j}$ based on $G_n$ and $\mathcal{P}_j$, and sample the corresponding RIR ${A}_{n,j}$. This process results in a dataset $\{(V_n, G_n, \{\mathcal{M}_{n,j}, \mathcal{A}_{n,j}\}^J)\}^N$ where $N=200$ and $J=100$ in our setup.

\vspace{-0.4cm}
\paragraph{Data Point Generation} %
To generate data points for our task, we select an observation $O_n$ and two random RIRs, a source ${A}_{n,S}$ and a target ${A}_{n,T}$ RIR, at location $l_n$. Here, $(V_n, G_n, {A}_{n,S})$ serves as the model input, $\mathcal{M}_{n,T}$ is the conditional target material mask, and ${A}_{n,T}$ is the target RIR to be generated. That is, for a specific input there could be multiple $\mathcal{M}_{n,T}$ with a corresponding ${A}_{n,T}$ to predict. This sampling strategy yields $\approx 1.68$ million unique data points in our dataset. See the Supp for a user study that analyze the perceptual differences between $A_S$ and $A_T$ in our dataset.

\vspace{-0.4cm}
\paragraph{Data Splits}
To evaluate model performance with respect to generalization, we define the following data splits. First, we divide the scenes into \emph{seen}, $\mathcal{S}_s$, and \emph{unseen} environments, $\mathcal{S}_u$. Additionally, we split the material profiles into \emph{seen} profiles $\mathcal{P}_s$, and \emph{unseen} ones $\mathcal{P}_u$. Further, we isolate a set of pairings between seen profiles, $\mathcal{K}=\{(\mathcal{P}_s\rightarrow \mathcal{P}_s)\}$ to serve as unseen mappings from source to target material configurations. 
The distinction between $\mathcal{P}_u$ and $\mathcal{K}$ lies in pairing configurations: $\mathcal{K}$ contains seen material profiles but with previously unseen source-target pairings, such as cases where walls are assigned to \emph{wood} or \emph{concrete} individually in training, but the model is not trained to anticipate transitions between these specific assignments. In contrast, $\mathcal{P}_u$ contains entirely unseen profiles for both source and target configurations, with the pairings also being unseen by definition. This setup allows for multiple evaluation scenarios of varying difficulty to test the model's generalization across scenes, profiles, and pairings. We use $\mathcal{S}_s$ and $\mathcal{P}_s$ for training, and $\mathcal{S}_u$, $\mathcal{P}_u$, $\mathcal{P}_s$, and $\mathcal{K}$ for evaluation (see Sec.~\ref{sec:eval}). %

\begin{figure*}[th!]
    \centering
    \includegraphics[width=\linewidth] {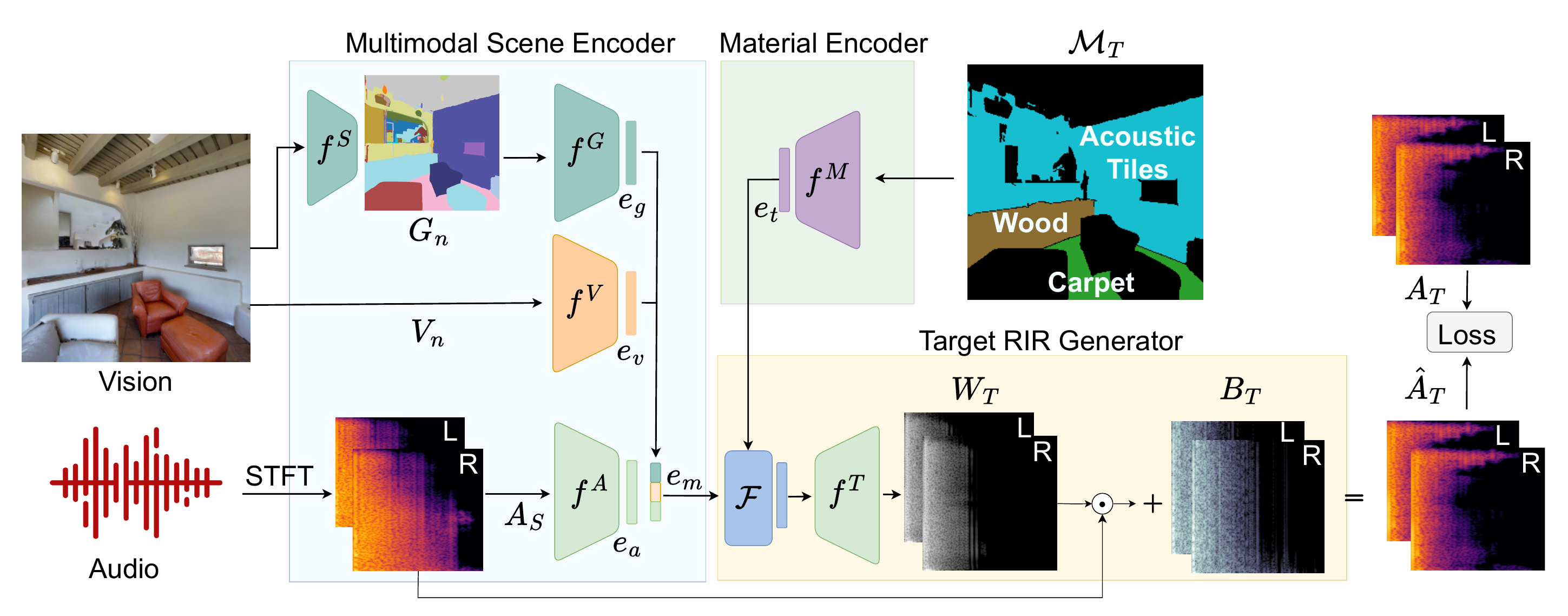}
    \caption{
        The architecture of our model. Given an audio-visual observation from the scene ($V_n$, $A_S$), the model encodes key visual and acoustic properties using a multimodal encoder. For a given arbitrary  target material assignment $\mathcal{M}_T$, the model then generates a weighting $W_T$ and a residual $B_T$ to adjust $A_S$ with a new reverberation pattern compatible with $\mathcal{M}_T$, thereby predicting the target RIR ($\hat{A}_T$). %
    }
    \label{fig:model}
    \vspace{-0.4cm}
\end{figure*}

\subsection{M-CAPA Model}
\label{sec:model}

We propose a novel approach for RIR prediction, conditioned on arbitrary material configurations within a given scene, named material-controlled acoustic profile anticipation (M-CAPA). Our model comprises three main components (see Fig.~\ref{fig:model}): 1) a multimodal scene encoder $f^E$, which processes visual input $V_n$, corresponding semantic segmentation mask $G_n$, and binaural echo response $A_{n,S}$ to create a multimodal embedding $e_m$ that captures both acoustic and visual properties of the scene; 2) a target material encoder $f^M$ that encodes the new material configuration of the scene into an embedding $e_t$; and 3) a conditional target RIR generator $f^T$, which uses both the scene encoding $e_m$ and target material information $e_t$ to predict changes in the target RIR $\hat{A}_{n,T}$ (for clarity we drop the sample index $n$ in the remainder of the text). We detail these components below.

\vspace{-0.4cm}
\paragraph{Multimodal Scene Encoder}
The model receives as input the RGB image $V_n \in \mathbb{R}^{H \times W \times 3}$ captured with a $90^\circ$ field-of-view (FoV) camera, and its associated semantic segmentation mask, $f^S(V_n) = G_n \in \mathbb{R}^{H \times W}$, where $f^S$ can be a pretrained semantic segmentation model,  %
and $H$ and $W$ are the height and width of the RGB. These images are each encoded via a four-layer convolutional UNet~\cite{unet} encoder block into a visual $e_v$ and a semantic $e_g$ embedding. 

The source binaural echo response RIR is first transformed into a binaural spectrogram magnitude image $A_S \in \mathbb{R}^{2 \times F \times T}$ using the short-time Fourier transform (STFT), where $F$ denotes the number of frequency bins and $T$ the number of overlapping time frames. This spectrogram is then encoded by a separate four-layer convolutional UNet encoder, $f^A$, yielding an acoustic embedding $e_a$. 

This combination of input modalities $(V, A_S)$ is advantageous because it avoids reliance on specialized hardware (e.g., a $360^\circ$ field-of-view camera) while still maintaining strong performance. This is due to the fact that echo responses inherently capture acoustic information from the entire room, including areas beyond the camera’s field of view.
Finally, the embeddings $e_v$, $e_g$, and $e_a$ are concatenated to form the multimodal scene embedding $e_m$.

\vspace{-0.4cm}
\paragraph{Target Material Encoder}
The arbitrary target material configuration of the scene is represented by a material segmentation mask $\mathcal{M}_T \in \mathbb{R}^{H \times W}$, where each element in $\mathcal{M}_T$ is a material class index $m_i \in M$. This mask can be defined by the user by assigning materials to objects and surfaces in $G_n$ or generated as part of the dataset during training (see Sec.~\ref{sec:dataset}). $\mathcal{M}_T$ represents a hypothetical new material configuration for which the user wishes to generate the RIR. The target material information is encoded with a convolutional encoder $f^M$, similar to $f^G$, into an embedding $e_t$.

\vspace{-0.4cm}
\paragraph{Material-Controlled RIR Generator}
With the scene information encoded in $e_m$ and the target material in $e_t$, we use both representations in a novel RIR prediction module, $f^T$, to generate $A_{T}$. This module first fuses the information from $e_m$ and $e_t$ using the fusion module $\mathcal{F}$, a convolution layer that combines the different modalities, and employs a decoder architecture with a series of four deconvolution layers, taking $\mathcal{F}(e_m, e_t)$ as input along with skip connections from $f^A$, to estimate a weighting mask $W_T \in \mathbb{R}^{2 \times F \times T}$ and a material residual information $B_T \in \mathbb{R}^{2 \times F \times T}$ such that:
\begin{equation}\label{eq:bias}
    \hat{A}_T = W_T \odot A_S + B_T,
\end{equation}
where $\odot$ is element wise multiplication.
The decoder $f^T$ predicts the target $\hat{A}_T$ by learning which parts of the input $A_S$ to emphasize or dampen using $W_T$ and which new reverberations to introduce using $B_T$, based on the conditional target material $\mathcal{M}_T$. We found that this novel formulation effectively anticipates changes in RIR, as new materials in the scene not only alter existing reverberation patterns but can also introduce reverberations in previously inactive frequency and time bins—a phenomenon not captured by conventional masking-based RIR prediction approaches (e.g.,~\cite{Adverb}), as we will demonstrate in our evaluation.

\vspace{-0.4cm}
\paragraph{Model Training}
Our model is trained end-to-end to minimize the error in the generated target RIR $\hat{A}_T$ compared to the ground truth $A_T$. The loss function is defined as:
\begin{equation}
    L_n = \lambda_1 ||\hat{A}_T - A_T||_2 + \lambda_2 ||\hat{A}_T - A_T||_1 + \lambda_3 L_D(\hat{A}_T, A_T),
\end{equation}
where $||\cdot||_2$ and $||\cdot||_1$ are the $L_2$ and $L_1$ losses based on the predicted $\hat{A}_T$ and ground truth $A_T$ binaural magnitude spectrograms. $L_D$ is an energy decay loss~\cite{majumder2022fsrir}, which aligns the temporal energy decay in the predicted RIR with the target, improving the quality of reverberations in $\hat{A}_T$. Based on validation performance, we set $\lambda_1 = \lambda_2 = 0.5$ and $\lambda_3 = 5\times 10^{-3}$.

\section{Experiments}
\label{sec:eval}

\begin{table*}[t]
    \centering
    \tiny
    \resizebox{1.\textwidth}{!}{
    \setlength{\tabcolsep}{2pt} %
    \renewcommand{\arraystretch}{1.} %
    \begin{tabular}{l|c c|r r r r|r r r r|r r r r} 
    \toprule
        & \multicolumn{2}{c|}{\textbf{Observation}} & \multicolumn{4}{c|}{\textbf{Seen Materials}} & \multicolumn{4}{c|}{\textbf{Unseen Materials}} & \multicolumn{4}{c}{\textbf{Unseen Pairings}} \\
    \textbf{Method} & \textbf{$A_s$} & \textbf{$V_n$} & \multicolumn{1}{c}{\textbf{L1}} & \multicolumn{1}{c}{\textbf{STFT}} & \multicolumn{1}{c}{\textbf{RTE}} & \multicolumn{1}{c|}{\textbf{CTE}} & \multicolumn{1}{c}{\textbf{L1}} & \multicolumn{1}{c}{\textbf{STFT}} & \multicolumn{1}{c}{\textbf{RTE}} & \multicolumn{1}{c|}{\textbf{CTE}} & \multicolumn{1}{c}{\textbf{L1}} & \multicolumn{1}{c}{\textbf{STFT}} & \multicolumn{1}{c}{\textbf{RTE}} & \multicolumn{1}{c}{\textbf{CTE}} \\   
    \midrule
    Direct Mapping    & \cmark  &   & 7.14 & 6.59 & 115.8 & 12.65 & 7.47 & 7.10 & 119.7 & 12.78 & 7.48 & 7.18 & 120.9 & 11.97  \\ 
    M-CAPA (Ours)     & \cmark  &   & {\bf 5.29} & {\bf 3.66} & {\bf 89.52} & {\bf 8.14} & {\bf 5.49} & {\bf 3.91} & {\bf 93.54} & {\bf 8.60} & {\bf 5.65} & {\bf 4.17} & {\bf 91.29} & {\bf 8.68} \\ %
       \midrule

     {Image2Reverb~\cite{image2reverb}} &      & {\cmark} & {14.68} & {7.89} & {245.16} & {18.76} & {14.13} & {7.59} & {223.36}  & {19.15} & {14.98} & {8.19} & {244.49} & {19.55} \\ %

     {FAST-RIR++~\cite{Fast-RIR,majumder2022fsrir}} & & {\cmark} & {16.73} & {25.06} & {317.18} & {21.47} & {14.81} & {28.39} & {231.83} & {16.83} & {16.41} & {31.02}  & {321.01} & {21.18} \\ %
    
    Material Agnostic                &      & \cmark & 8.95 & 11.16 & 121.43 & 12.21 & 9.21 & 11.65 & 122.7 & 13.66 & 9.41 & 11.93 & 124.75 & 14.19 \\
    Material Aware                   &      & \cmark & 8.91 & 11.19 & 98.02 & 11.48 & 8.91 & 11.29 & 98.06 & 11.75 & 9.21 & 11.52 & 98.72 & 11.19 \\ 
   
    M-CAPA (Ours) &  & \cmark & \textbf{5.92} & \textbf{5.49} & \textbf{89.23} & \textbf{8.41} &\textbf{6.06}  & {\bf 5.76} &\textbf{92.80} &\textbf{9.05} &\textbf{6.30}& {\bf 6.17} & \textbf{91.75}& \textbf{8.95} \\  %
     \midrule
   
    AV-RIR~\cite{AV-RIR}    & \cmark  & \cmark      & 7.31 & 6.65 & 99.34 & 10.92 & 7.59 & 7.17 & 99.10 & 11.35 & 7.67 & 7.25 & 98.46 & 10.56 \\ 
    M-CAPA (Ours)           & \cmark  & \cmark      & {\bf 5.10} & {\bf 3.61} & \textbf{87.49} & \textbf{7.98} & {\bf 5.27} & {\bf 3.87} & {\bf 91.44} & {\bf 8.44} & {\bf 5.46} & {\bf 4.15} & \textbf{90.77} & {\bf 8.56} \\

    \bottomrule
    \end{tabular}
    }
    \caption{
        Results on unseen environments for our three test splits: $D_{us}$ with seen material profiles, $D_{uu}$ with unseen material profiles, and $D_{uk}$ with unseen profile pairings. STFT and $L_1$ are scaled by $\times 10^{-2}$, RTE is in milliseconds (ms), and CTE in decibels (dB). Lower values indicate better performance for all metrics. We group the models based on the input modalities: audio-only (top), vision-only (middle), and audio-visual (bottom). Our model outperforms all baselines across these groups and all metrics.
    }

    \vspace{-0.4cm}
    \label{tab:main}
\end{table*}
We evaluate our model's performance on RIR generation using the Acoustic Wonderland Dataset (AcWon) (Sec.~\ref{sec:dataset}) and compare it with several state-of-the-art (SoTA) methods and baselines to demonstrate the effectiveness of our approach (Sec.~\ref{sec:results}). We provide a detailed analysis of our model in Sec.~\ref{sec:analysis}. Next, we outline our evaluation setup, with more details provided in the Supp.

\vspace{-0.4cm}
\paragraph{Implementation Details}
For RGB images, $V_n$, we use a resolution of $256\times256$ and sample the binaural echo response RIRs, $A$, from SSv2 at a rate of $16$kHz and a duration of $0.5$ seconds. Spectrograms are generated using STFT with a Hann window ~\cite{1455106} of length 256, hop length of 32, and FFT size of 511, resulting in a binaural spectrogram with dimensions $2\times256\times256$. Additionally, we extract the semantic segmentation mask $G_n$ from SSv2 and also test with an inferred $G_n$ from a pretrained model~\cite{yang2022bevformer}. Our model is trained on a single GPU using the Adam optimizer~\cite{adam-optimizer} with a learning rate of $10^{-3}$ and a batch size of $64$. %

\vspace{-0.4cm}
\paragraph{Dataset Splits}
We use the AcWon dataset and split the $84$ MP3D scenes into $|\mathcal{S}_s|=76$ \emph{seen} and $|\mathcal{S}_u|=8$ \emph{unseen} environments. The $2,673$ material profiles are split into $|\mathcal{P}_s|=2,405$ \emph{seen} and $|\mathcal{P}_u|=268$ \emph{unseen} profiles. Furthermore, we isolate $|\mathcal{K}|=2000$ source-to-target material profile mappings to be used exclusively for evaluation, not for training. 
Our training data consists of $D^{tr}=\{\mathcal{S}_s, \mathcal{P}_s\}$, and we create three evaluation splits: $D_{us}=\{S_u, \mathcal{P}_s\}$, $D_{uu}=\{\mathcal{S}_u, \mathcal{P}_u\}$, and $D_{uk}=\{\mathcal{S}_u, \mathcal{K}\}$. For validation splits $D^v$, we follow similar criteria as the previous three, using three of the $\mathcal{S}_u$ scenes and reserving the remainder for testing as $D^t$. The test set comprises $6,000$ samples, with $2,000$ samples each for $D^t_{us}$, $D^t_{uu}$, and $D^t_{uk}$.

\vspace{-0.3cm}
\paragraph{SoTA Methods and Baselines}
We compare our M-CAPA model against the following SoTA methods and baselines (see Supp for more details):
\begin{itemize}[leftmargin=*,label={},align=left]
    \item \textbf{Direct Mapping $A_S\rightarrow A_T$}: This baseline outputs $A_S$ as the predicted $\hat{A}_T$, capturing the scene's mean acoustic characteristics under the original material configuration. This helps quantify improvements achieved by our model in predicting $A_T$ conditioned on the target material $\mathcal{M}_T$.
    \item \textbf{Material Agnostic Matcher}: It finds the closest visual match from the training set based on similarity of the visual embedding $e_v$ and retrieves an RIR associated with that location, $l_n$, as the output. It serves  as a representative of models that memorize training RIRs and predict based on visual similarity between the test and training scenes.
    \item \textbf{Material Aware Matcher}: Similar to the previous baseline, but in addition to visual similarity, it also considers the similarity of material distributions. It retrieves an RIR based on both visual and material similarity between the test sample and training data.
    \item \textbf{Image2Reverb}~\cite{image2reverb}: A vision-only RIR prediction SoTA model, which uses RGB and depth maps to predict the RIR of the input scene. We train the model on our training split using the code provided by the authors. %
    \item \textbf{FAST-RIR++}~\cite{Fast-RIR,majumder2022fsrir}: Fast-RIR~\cite{Fast-RIR} is a GAN-based SoTA approach that uses the scene properties to synthesize RIRs for rectangular rooms. We follow  the improved version introduced by~\cite{majumder2022fsrir} and use the estimated RT60 and DRR from $A_S$, and GT depth maps as  inputs to the model. %
    \item \textbf{AV-RIR}~\cite{AV-RIR}: A SoTA audio-visual model with explicit material modeling for RIR prediction. Instead of inferring  source RIR from reverberant speech, we adapt this model to our setting by providing $A_S$ directly. We replace the late components of the RIR by retrieving the closest training sample based on  target material similarity  to generate $\hat{A}_T$. 
\end{itemize}

\vspace{-0.3cm}
\paragraph{Metrics}
We evaluate performance using standard RIR prediction metrics: 1) \textbf{STFT Error}: the mean squared error between predicted and target RIR based on the magnitude spectrograms; 2) \textbf{L1 Distance}: similar to STFT, but measures $L_1$ distance; 3) \textbf{RT60 Error (RTE)}~\cite{ratnarajah2021ts}: the error in RT60 values of the predicted RIR, and 4) \textbf{Early-to-Late Index Error (CTE)}~\cite{ratnarajah2021ts}: capturing the error in the ratio of early- to late-sound energy received. 
STFT and $L_1$ metrics capture fine-grained prediction errors, while RTE and CTE focus on acoustic and reverberation characteristics.

\subsection{Target RIR Generation Results}\label{sec:results}

In Table~\ref{tab:main}, we present the performance of our model (M-CAPA) in comparison to existing methods and baseline approaches on the three splits of the test dataset $D^t$. For a fair comparison, we evaluate three versions of our model, each using different input modalities to match the corresponding baselines. We observe that, in general, audio-only models outperform those that rely solely on vision, while audio-visual methods achieve the best performance.

For retrieval-based RIR predictors, we find that the material-aware baseline, which considers the target material distribution, outperforms the material-agnostic method, FAST-RIR++~\cite{Fast-RIR,majumder2022fsrir}, and Image2Reverb~\cite{image2reverb}  (except for STFT). %
Furthermore, AV-RIR~\cite{AV-RIR} improves upon these retrieval methods by leveraging the estimated  source RIR of the original scene, $A_S$, and transferring late reverberation patterns from a retrieved RIR with a similar visual and material distribution within the training data. Nevertheless,  while AV-RIR improves the RTE and CTE performance, all other methods still struggle to surpass the simple direct mapping baseline. This may be due to the challenges in effectively modeling the impact of material properties on the target RIR and the substantial differences between the seen scenes $\mathcal{S}_s$ used for training and the unseen scenes $\mathcal{S}_u$ used for testing, which require strong generalization capabilities.

Our approach outperforms all baselines and methods across the various input modalities, metrics, and testing setups, demonstrating the robustness and effectiveness of our model. Interestingly, the vision-only variant of our model (using $V_n$ and without $A_S$ or $G_n$) still outperforms all competing methods, including those with audio-visual inputs. This demonstrates that while audio observations are beneficial, strong performance can still be achieved using vision alone, simplifying the input requirements. Across different splits, performance on $D^t_{uu}$ and $D^t_{uk}$ is lower than on $D^t_{us}$. This is expected, as these settings require the model to generalize not only to unseen scenes but also to unseen material configurations and profile pairings. Analyzing the performance of different models on separate splits using the seen scenes $\mathcal{S}_s$ (see Supp for details) further highlights that generalization across scenes remains a major factor contributing to prediction errors across all methods.

\begin{table}[t]
    \centering
    \resizebox{1.\columnwidth}{!}{
    \setlength{\tabcolsep}{2pt} %
    \renewcommand{\arraystretch}{1.} %
    \begin{tabular}{l|r r r r} 
    \toprule
        \textbf{Method} & \textbf{L1} & \textbf{STFT} & \textbf{RTE} & \textbf{CTE}\\
    \midrule
    M-CAPA (audio-visual)               & 5.27 & 3.87 & 91.44 & 8.44 \\ 
    \midrule
    
    a) Ours w/o $\mathcal{M}_T$ & 5.61 & 4.06 & 109.46 & 9.19 \\ 
    b) Ours w/o $B_T$           & 5.75 & 4.93 & 105.19 & 10.83 \\ 
    c) Ours w/ Inferred $G_n$   & 5.63 & 3.99 & 97.63 & 9.10 \\ 
    d) Ours w/ Changed $\mathcal{M}_T$  & 5.47 & 4.00 & 96.36 & 9.04 \\ 
    \bottomrule
    \end{tabular}
    }
    \caption{Ablation of our model on the test split $D_{uu}$. Lower is better for all metrics. For the other splits, see Supp.
    }
    \vspace{-0.5cm}
    \label{tab:ablations}
\end{table}

\subsection{Model Analysis}\label{sec:analysis}

\paragraph{Ablations}
Table~\ref{tab:ablations} presents various ablations of our model to investigate the contribution of different components. 

First, excluding the target material information (row $a$) negatively impacts performance, especially in metrics that capture key RIR acoustic properties, such as RTE and CTE. We also evaluate our novel formulation for RIR generation (Eq.~\ref{eq:bias}), finding that not explicitly modeling the novel impact, $B_T$, of the target material on $A_T$ leads to weaker performance. Notably, learning only masking weights is not sufficient for precise predictions (row $b$).

Furthermore,  in row $c$, we test the effect of using a pretrained semantic segmentation model~\cite{yang2022bevformer} to infer $G_n$ from $V_n$, rather than retrieving $G_n$ directly from SSv2. This leads to  a small drop in performance, suggesting a gap that could be mitigated with a more effective segmentation model.  

Finally, in row $d$, we examine whether it is necessary to provide a full material assignment in $\mathcal{M}_T$ or if specifying only the changed materials is sufficient. Our results indicate that a complete target material mask,  while it helps, it is not necessary, which simplifies the input requirements for our model. See the Supp for loss ablations  and comparisons of the computational cost between our model and the baselines. %

\begin{figure}[t]
    \centering
    \begin{subfigure}[t]{0.49\columnwidth}
        \centering
        \includegraphics[width=\linewidth]{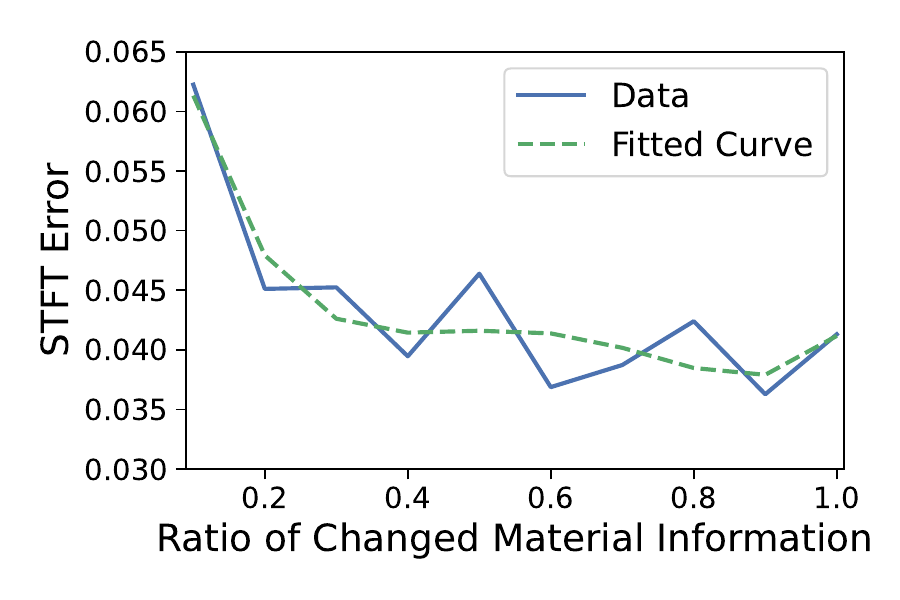} 
        \caption{}
        \label{fig:area_stft_duu}
    \end{subfigure}%
    \hfill
    \begin{subfigure}[t]{0.49\columnwidth}
        \centering
        \includegraphics[width=\linewidth]{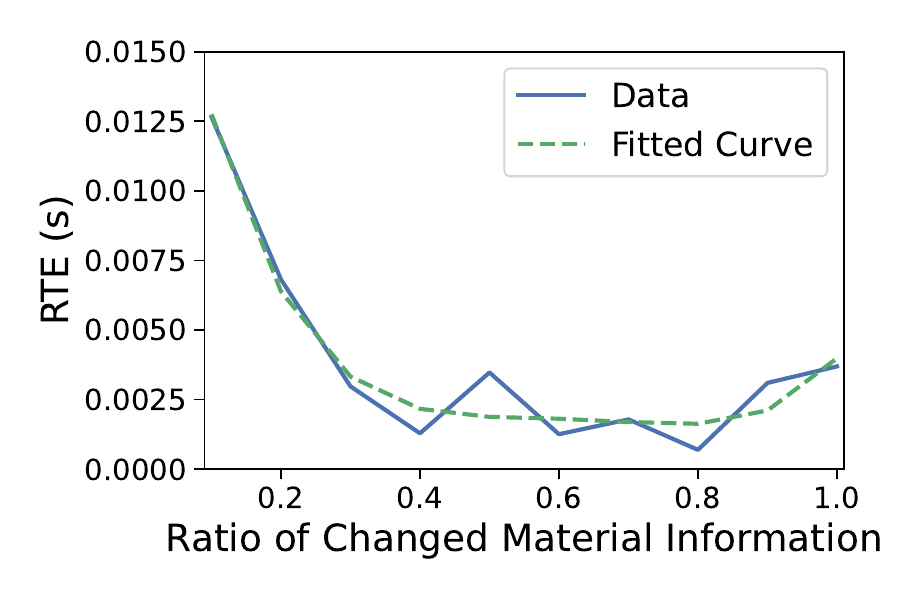} 
        \caption{}
        \label{fig:area_rte_duu}
    \end{subfigure}
    \\
    \begin{subfigure}[t]{0.49\columnwidth}
        \centering
        \includegraphics[width=\linewidth]{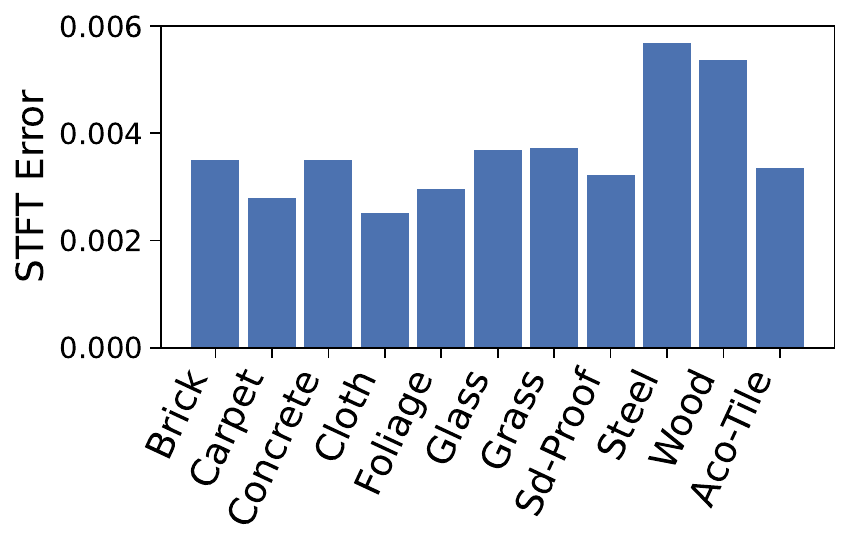} 
        \caption{}
        \label{fig:mat_stft_duu}
    \end{subfigure}%
    \hfill
    \begin{subfigure}[t]{0.49\columnwidth}
        \centering
        \includegraphics[width=\linewidth]{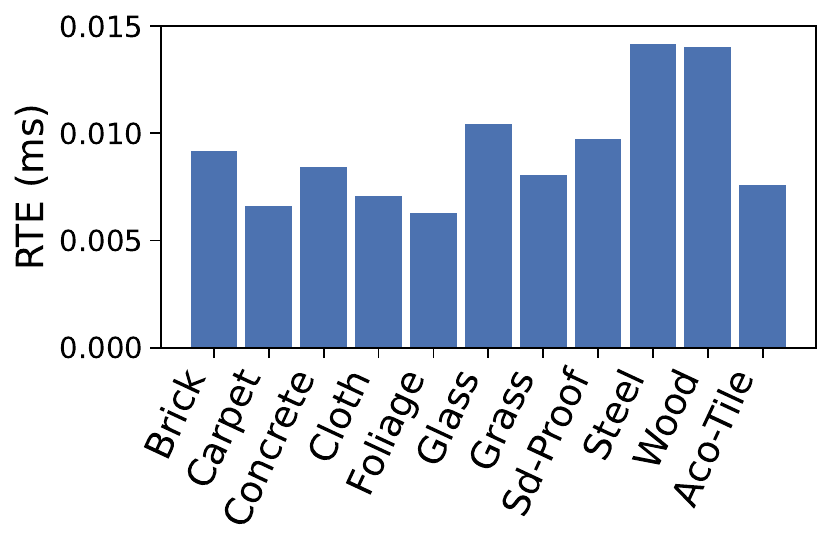} 
        \caption{}
        \label{fig:mat_rte_duu}
    \end{subfigure}
    \vspace{-0.2cm}
    \caption{
        Performance analysis of our model with respect to the percentage of new material assignments in $\mathcal{M}_T$ (a and b) and across different material classes (c and d).
    }
    \label{fig:error}
    \vspace{-0.5cm}
\end{figure}

\vspace{-0.4cm}
\paragraph{Performance Analysis} %
In Fig.~\ref{fig:error}, we present a detailed analysis of our model's performance on the $D_{uu}$ test split. First, we examine the correlation between the relative area size associated with new material assignments in $\mathcal{M}_T$ and the performance metrics. As shown in Fig.~\ref{fig:area_stft_duu} and Fig.~\ref{fig:area_rte_duu}, an interesting relationship exists between these variables. When the modified area in $\mathcal{M}_T$ is small, we observe a relatively large error, which decreases as the material assignments cover a larger portion of $\mathcal{M}_T$. This may be due to the difficulty in capturing the impact of small material changes in the scene (e.g., changing the material of a chair) on the final RIR. Additionally, smaller objects often have irregular shapes, which makes predicting how they interact with sound in the target RIR more challenging.

\begin{figure*}[t]
    \vspace{-0.45cm}
    \centering
    \begin{subfigure}[t]{0.75\textwidth}
        \centering
        \includegraphics[width=\linewidth]{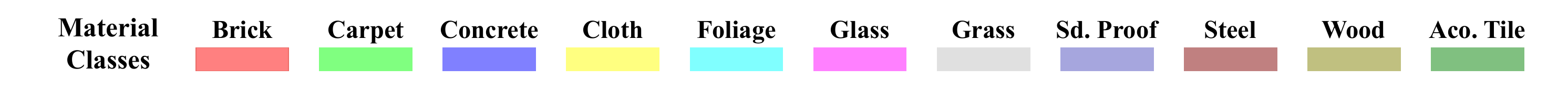} 
    \end{subfigure}\\
    \vspace{-0.03cm}
    \begin{subfigure}[t]{0.45\textwidth}
        \centering
        \includegraphics[width=\linewidth]{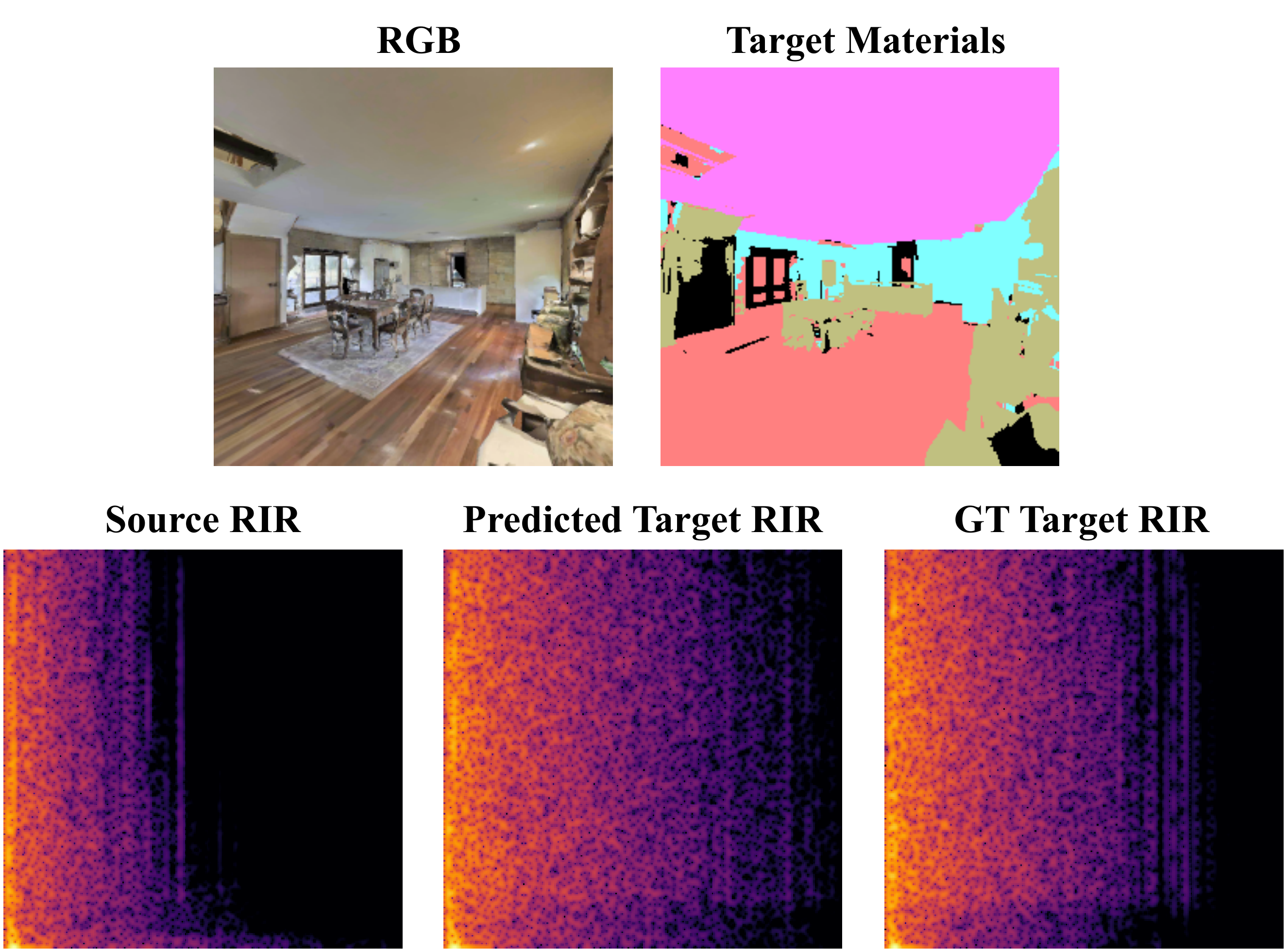} 
        \caption{}
        \label{fig:qual_1}
    \end{subfigure}%
    \hfill
    \begin{subfigure}[t]{0.45\textwidth}
        \centering
        \includegraphics[width=\linewidth]{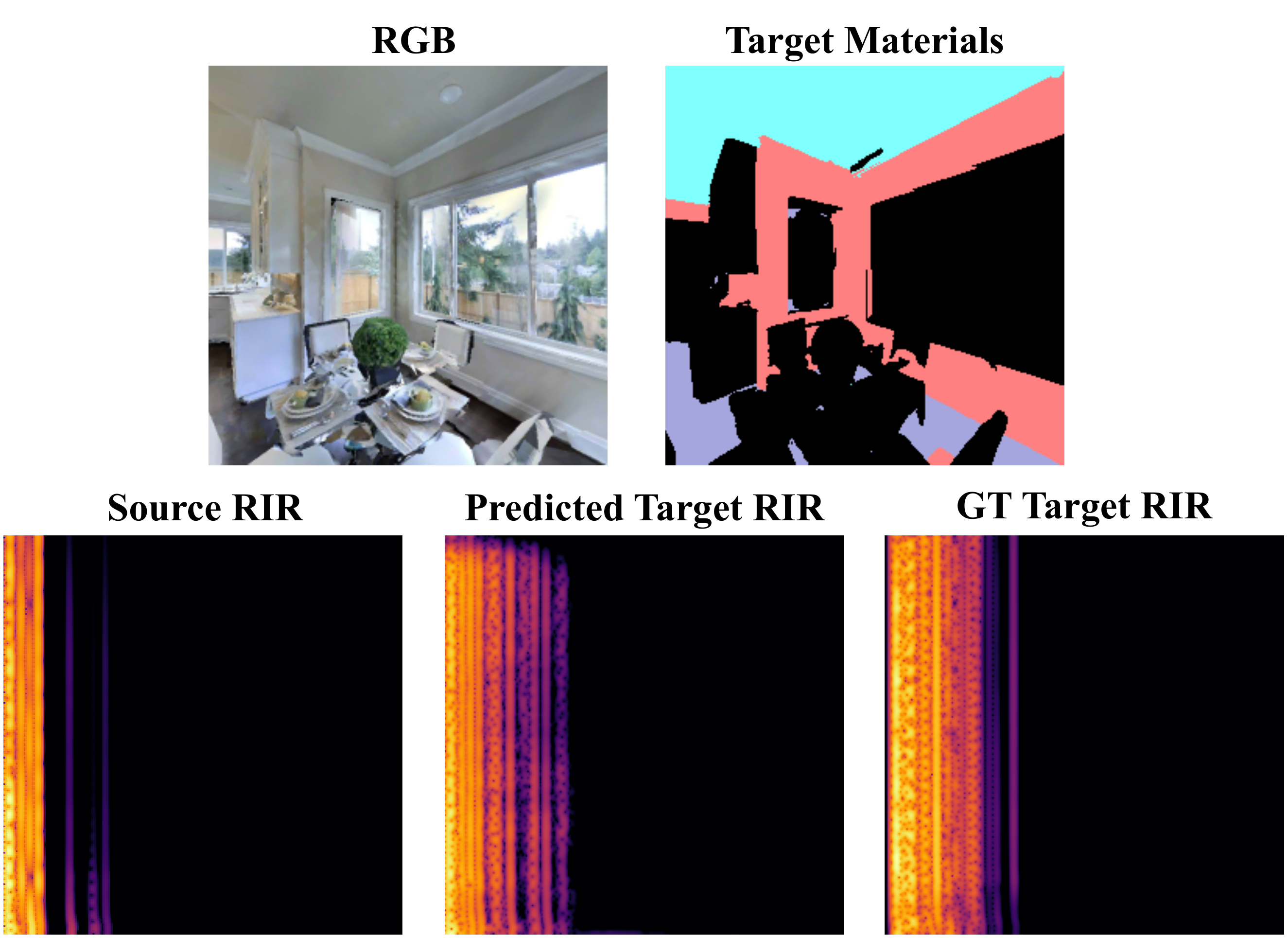} 
        \caption{}
        \label{fig:qual_2}
    \end{subfigure}
    \caption{
        Qualitative results. Our model effectively captures the impact of target material configurations on the generated target RIR, even when these patterns are novel and absent from the source RIR (a and b). For brevity, we only show one channel of the binaural RIRs. %
    }
    \label{fig:qual}
    \vspace{-0.5cm}
\end{figure*}

The lowest error is observed when the changed material covers between $50\%$ and $70\%$ of the mask, typically corresponding to objects like walls, floors, and ceilings. These surfaces tend to be flat and regular, which makes it relatively easier to model their acoustic effects. Interestingly, the error increases slightly when new material assignments cover almost the entire scene.

We further analyze performance across different material classes in Fig.~\ref{fig:mat_stft_duu} and Fig.~\ref{fig:mat_rte_duu}. Our model appears to exhibit higher error rates with material classes such as \emph{wood} and \emph{steel}, compared to \emph{cloth}, \emph{foliage}, and \emph{acoustic tiles}, which seem easier for our model to handle. This difference could be due to the intrinsic properties of these materials (i.e, how they absorb, reflect, and cause reverberations) across various frequency bands. Additional analysis is needed to better understand these material-specific behaviors.

\vspace{-0.4cm}
\paragraph{Qualitative Results}
In Fig.~\ref{fig:qual}, we present two qualitative results from our model. Our model effectively incorporates changes in the target material mask and simulates their impact on the predicted target RIR of the scene. For instance, the model successfully introduces new reverberation patterns to reflect the effect of assigning a \emph{brick} material to the floor (Fig.~\ref{fig:qual_1}) or \emph{foliage} to the ceiling (Fig.~\ref{fig:qual_2}). Note that these reverberation patterns were absent in the source echo response RIR; nonetheless, our model accurately captures the effects of the changed materials in the output. Please refer to the supplementary video to experience the impact of material changes on the generated  target RIRs. %

\begin{figure}[t]
    \centering
        \includegraphics[width=0.9\columnwidth]{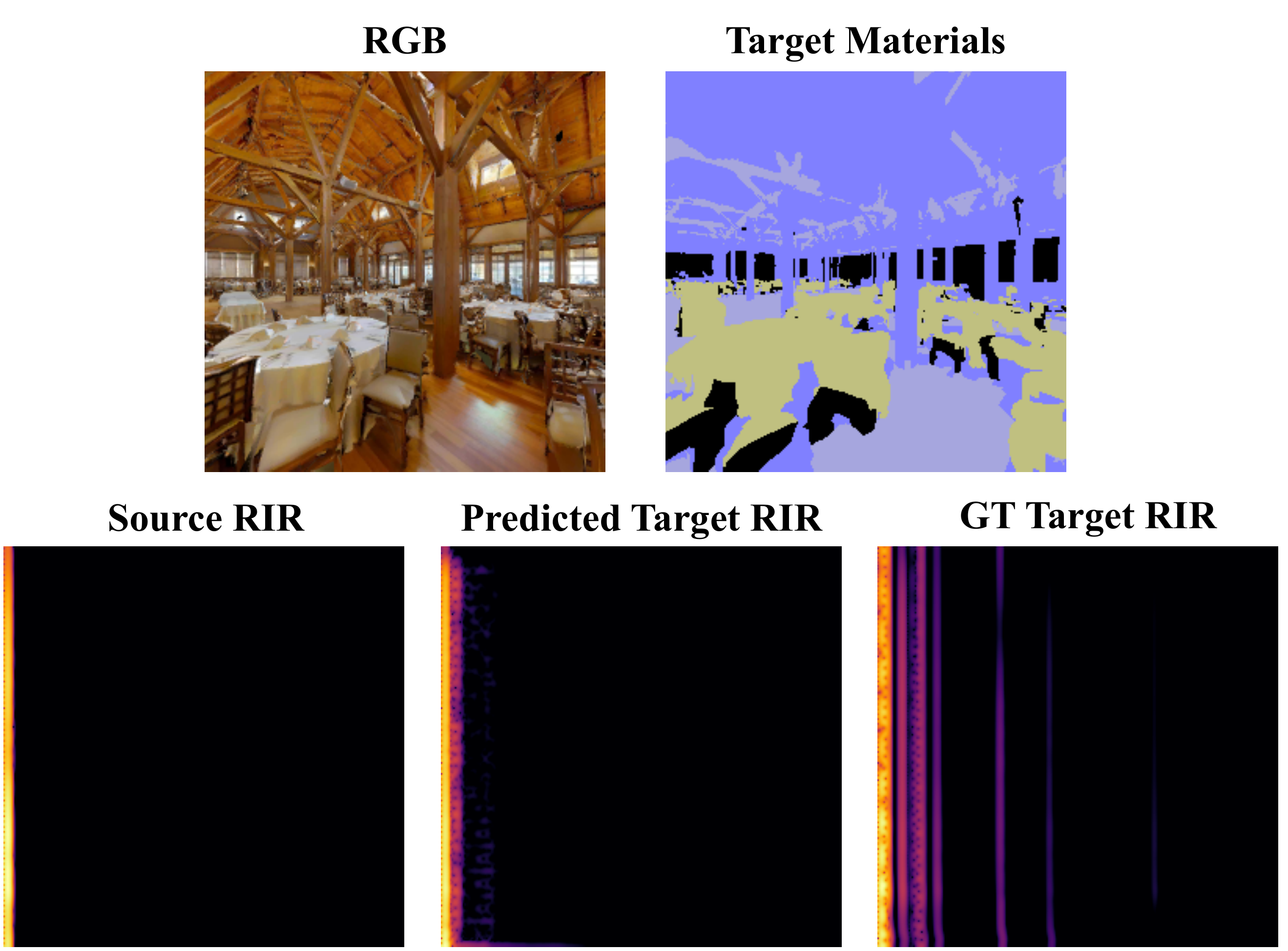} 
    \caption{For scenes with complex and highly irregular shapes, such as the ceiling in this example, the model encounters challenges in accurately estimating the target RIR.}
    \label{fig:failure}
    \vspace{-0.5cm}
\end{figure}

\vspace{-0.4cm}
\paragraph{Failure Cases and Limitations} %
While our model performs well overall, we observed some cases where the proposed approach encounters difficulties. Specifically, changing the materials of irregularly shaped objects often leads to suboptimal estimates of the target RIR. We provide an example of such a scenario in Fig.~\ref{fig:failure}. In this example, the top of the scene comprises an intricate set of columns, domes, and beams, along with a dense arrangement of chairs and tables. When changing the material of the top area to \emph{concrete}, the model struggled to accurately capture this change, likely due to the strong irregularity in the ceiling's shape. 

Furthermore, when analyzing the impact of acoustic noise on the robustness of our predictions (see Supp for details), we find that performance degrades as noise levels increase, due to the reduced quality of the source echo response RIR. However, we anticipate that training the model with acoustic augmentation techniques with noisy inputs could improve the approach's robustness. 
Lastly, our approach does not currently account for the introduction of new, unseen material classes at inference time. Addressing this limitation is an interesting direction for future work.

\subsection{User Study on Real-World Data}\label{sec:eval_users} 

Since there is no real-world dataset compatible with our task, we collected samples (RGBs) from two real scenes. In each scene, we assigned the target materials to one of three classes (Carpet, Brick, and Glass) and used our vision-only model to generate the target RIR. After a brief training with simulated data,  5 users were asked to identify the target material based solely on speech convolved with the predicted $A_T$. The overall accuracy was 61.1\% (random chance: 33\%), demonstrating that our model effectively encodes target material properties and generalizes to real-world data. See Supp and video for details.

\section{Conclusion}
\label{sec:conclusion}
This work introduces a novel task and an approach for dynamically controlling the generation of a target Room Impulse Response (RIR) using arbitrary material configurations at inference time. Additionally, we have compiled a new dataset, the Acoustic Wonderland dataset, designed to support the development and evaluation of multimodal methods for material-aware acoustic profile modeling within a 3D scene. We anticipate that the proposed task and dataset will be of significant interest to the research community, and enable new applications in AR/VR, creative design, sound engineering, and spatial planning.

{
    \small
    \bibliographystyle{ieeenat_fullname}
    \bibliography{main}
}
\clearpage

\section{Supplementary Material}
\label{sec:supp}
In this supplementary material, we provide further details about: 
\begin{itemize} 
    \item Supplementary video (with audio) Sec.~\ref{sec:supp_video} for qualitative evaluation of our model predictions as stated in Sec.~\ref{sec:eval}.
    \item Real-World Generalization (Sec. ~\ref{sec:real-world}) as mentioned in Sec.~\ref{sec:eval}.
    \item Ablations on other test splits (Sec.~\ref{sec:supp_eval_abl}) as mentioned in Sec.~\ref{sec:eval}, Table~\ref{tab:ablations}. 
    \item Loss ablations Sec.~\ref{sec:loss-ablations}  and computational cost analysis Sec.~\ref{sec:supp_compcost} as stated in Sec.~\ref{sec:eval}.
    \item Performance analysis on other test splits (Sec.~\ref{sec:supp_perf}) as stated in Sec.~\ref{sec:eval}. 
    \item Evaluation results on seen splits (Sec.~\ref{sec:supp_eval_seen}) as stated in (Sec.~\ref{sec:eval}).
    \item Robustness to noise experiments (Sec.~\ref{sec:supp_noise}) as noted in Sec.~\ref{sec:eval}. 
    \item Acoustic Wonderland dataset (Sec.~\ref{sec:supp_datasest}), as mentioned in Sec.~\ref{sec:dataset}, and a user study on the perceptual differences as mentioned in Sec.~\ref{sec:dataset}.
    \item Model Architecture details (Sec.~\ref{sec:supp_model}).
    \item Evaluation setup (Sec.~\ref{sec:supp_eval_setup}), as mentioned in Sec.~\ref{sec:eval}. 
\end{itemize}

\subsection{Supplementary Video}\label{sec:supp_video} 

We provide a supplementary video,  see the project page, to illustrate the qualitative results produced by our model, M-CAPA. The video begins with a brief overview of the motivation and contributions of this work. It then presents qualitative results by showcasing a variety of speech sounds from the datasets~\cite{richey2018voices} and~\cite{librispeech}, convolved with the predicted target room impulse response (RIR), $\hat{A}_T$. These examples emphasize the quality of the predictions and demonstrate how effectively the model captures the diverse target material configurations introduced in the input scenes. 

Furthermore, the video highlights failure cases where the model encountered difficulties in accurately representing material changes, thereby shedding light on challenges that remain to be addressed. For instance, M-CAPA struggles to model environmental acoustics when significant material changes are applied to large objects with highly irregular shapes. Additionally, we observe suboptimal performance when certain materials, such as \textit{Sound-Proof} and \textit{Steel}, are extensively used in the target material mask.

\subsection{Ablations On Other Test Splits}\label{sec:supp_eval_abl}

We present ablation results on the remaining test splits, $D_{us}$ and $D_{uk}$, in Table~\ref{tab:ablations-remaining}. Similar trends to those reported in Table~\ref{tab:ablations} in the main text are observed. Our complete model, \emph{M-CAPA}, achieves the best overall performance across all splits. 
Notably, as shown in \textit{row b}, incorporating $B_T$ allows the model to learn the differences between $A_S$ and $A_T$ that arise from selecting target materials, which introduce new types of reverberations not present in $A_S$. This incorporation enhances learning, particularly for acoustic metrics such as RTE and CTE. Furthermore, in \textit{row a}, excluding the target material change and relying solely on visual cues and $A_S$ to predict $A_T$ leads to a noticeable degradation in performance.

\begin{table}[t]
    \centering
    \tiny
    \resizebox{\columnwidth}{!}{
    \setlength{\tabcolsep}{2pt} %
    \renewcommand{\arraystretch}{1.} %
    \begin{tabular}{l|r r r r|r r r r} 
    \toprule
    & \multicolumn{8}{c}{\textbf{Unseen Environments}}\\
    & \multicolumn{4}{c|}{\textbf{Seen Materials}} & \multicolumn{4}{c}{\textbf{Unseen Pairings}} \\
    \textbf{Method} & \multicolumn{1}{c}{\textbf{L1}} & \multicolumn{1}{c}{\textbf{STFT}} & \multicolumn{1}{c}{\textbf{RTE}} & \multicolumn{1}{c|}{\textbf{CTE}} & \multicolumn{1}{c}{\textbf{L1}} & \multicolumn{1}{c}{\textbf{STFT}} & \multicolumn{1}{c}{\textbf{RTE}} & \multicolumn{1}{c}{\textbf{CTE}} \\      %
    \midrule
    M-CAPA (Ours)                   & 5.10	& 3.62	&88.15&	8.04   & 5.47 & 4.15 & 91.32 & 8.57\\
    \midrule
    
    a) Ours w/o $\mathcal{M}_T$  &  5.39 &	3.78 &	104.77 &	8.67 &	5.77 &	4.35 &	107.53& 9.13 \\ 
    b) Ours w/o $B_T$            & 5.52	&4.52&	98.30	&10.79&	5.93	&5.17&	104.72&	10.51\\ 
    c) Ours w/ Inferred $G_n$    & 5.42	& 3.72	& 98.46 &	8.53&	5.79&	4.27&	99.70&	9.03\\ 
    d) Ours w/ Changed $\mathcal{M}_T$   & 5.27	& 3.74 &	94.97&	8.48	&5.63	&4.29	&96.81&	8.95\\ 
    
    \bottomrule
    \end{tabular}
    }
    \caption{
        Ablation results of our model on unseen environments using test sets $D_{us}$ (seen material profiles) and $D_{uk}$ (unseen material profile pairings). The results exhibit similar trends to those observed on $D_{uu}$. For all metrics, lower values indicate better performance.
    }

    \label{tab:ablations-remaining}
\end{table}

\subsection{Real-World Generalization}\label{sec:real-world}

\begin{figure}[t]
    \centering
    \includegraphics[width=1\linewidth]{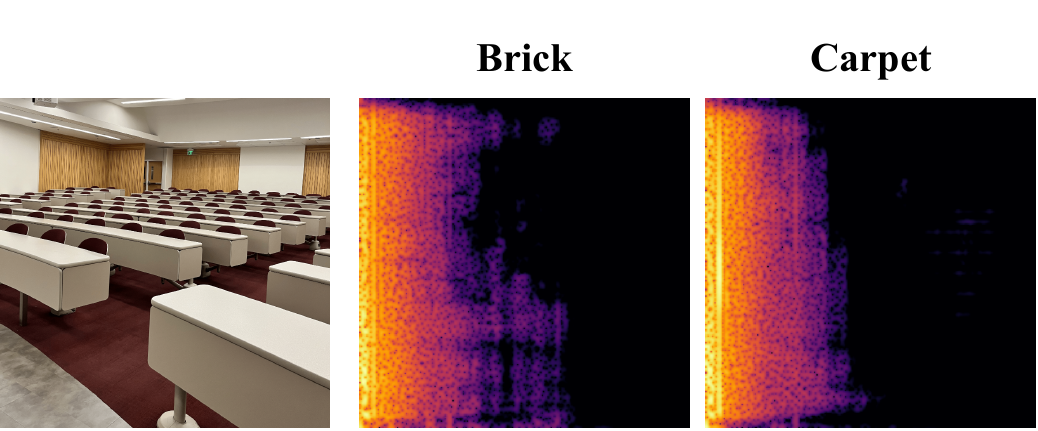}
    \caption{
        Predicted RIRs from vision-only M-CAPA in an auditorium classroom environment where $M_T$=Brick and $M_T$=Carpet  
    }
    \label{fig:real-world-qual}
\end{figure}

To asses M-CAPA's ability to generalize to real-world samples,
we collected RGB images from two real-world scenes and used our vision-only M-CAPA to generate a target RIR ($A_T$). The target material of the objects in the scenes was set to one of three classes \textit{carpet, brick, and glass} (Figure~\ref{fig:real-world-qual} shows qualitative results).

\begin{figure}[t]
    \centering
    \begin{subfigure}[t]{0.9\columnwidth}
        \centering
        \includegraphics[width=\linewidth]{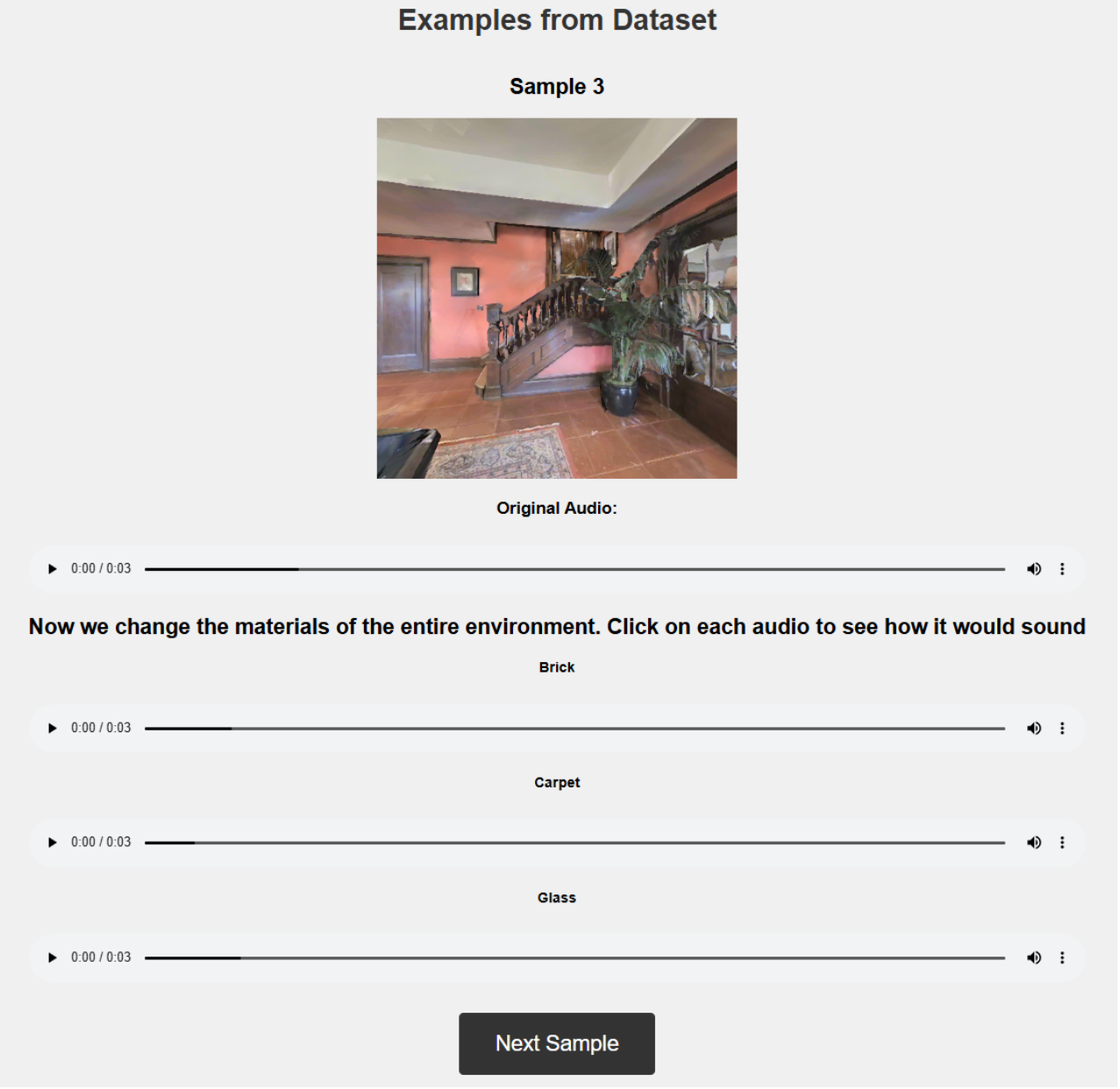} 
        \caption{}
        \label{fig:user-study-training}
    \end{subfigure}
    
    \vspace{0.3cm} %
    
    \begin{subfigure}[t]{0.9\columnwidth}
        \centering
        \includegraphics[width=\linewidth]{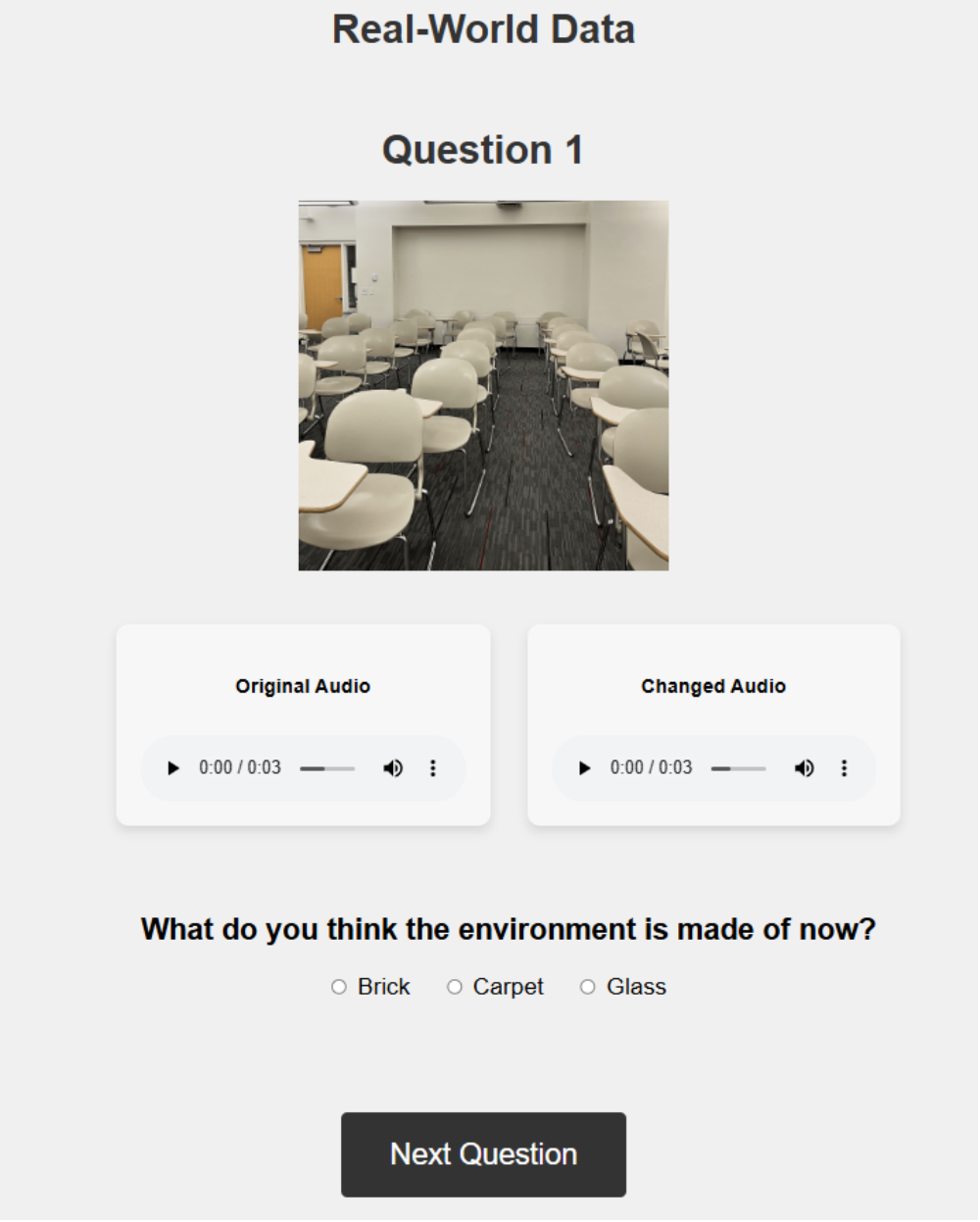} 
        \caption{}
        \label{fig:user-study-question}
    \end{subfigure}

    \vspace{-0.2cm}
    \caption{ 
        User interface for the real-world user study. a) Interface for user training b) Interface for the real-world samples.
    }
    \label{fig:user-study-interface}
    \vspace{-0.5cm}
\end{figure}

Then, we conduct a user study (\ref{sec:eval_users}) to measure M-CAPA's performance. 
We ask 5 users to go through a brief training so they may distinguish the acoustic properties of different materials (Figure~\ref{fig:user-study-training}). Afterwards, we ask them to listen to the predictions by M-CAPA on the real-world samples when $A_T$ is convolved with speech, and ask them to identify the target material used to generate $A_T$ as one of the three materials: Brick, Carpet, and Glass (Figure~\ref{fig:user-study-question}). Overall, the accuracy achieved by the users in identifying the correct material in this task was 61.1\% (random chance: 33\%), showing that our model successfully encodes the target material signature in $A_T$ even in samples from real-world scenes.

\subsection{Loss Ablations}\label{sec:loss-ablations}
As discussed in Sec.\ref{sec:model}, our model is trained with $L1$, $L2$ and energy decay loss \cite{majumder2022fsrir}. We investigate the impact of each loss as our learning objective by performing ablations on the losses (Table \ref{tab:loss-ablations-updated}). We see from  row (a) and row (b) that $L1$ is the most important loss in minimizing error between predicted RIR and ground truth RIR. However, $L2$ plays a vital role in ensuring that the STFT loss is minimized, and that loss between acoustic parameters is consequently reduced. The energy decay loss acts as supervision for the acoustic metrics, CTE and RTE, ensuring that the reverberation time and early-to-late reflections of the predicted RIR are aligned with the ground truth RIR.

\begin{table}[t]
    \centering
    \resizebox{1.\columnwidth}{!}{
    \setlength{\tabcolsep}{2pt} %
    \renewcommand{\arraystretch}{1.} %
    \begin{tabular}{l|r r r r} 
    \toprule
        \textbf{Loss} & \textbf{L1} & \textbf{STFT} & \textbf{RTE} & \textbf{CTE}\\
    \midrule
    $L_1+L_2+$Energy Decay              & 5.29 & 3.87 & 90.61 & 8.52 \\  
    \midrule
    a) $L_1$ Only           & 5.46  & 4.13  & 97.92 & 9.47\\ 
    b) $L_2$ Only     & 6.19  & 4.00  & 241.41 & 9.22\\
    c) $L_1+$Energy Decay     & 5.55 & 4.15 & 99.00 & 9.45\\
    d) $L_2+$Energy Decay     & 6.47  & 4.12  & 248.69 & 9.12\\
    e) $L_1+L_2$     & 5.59 & 3.99 & 109.27 &  9.26\\
    \bottomrule
    \end{tabular}
    }
    \caption{Ablation of losses
    }
    \vspace{-0.5cm}
    \label{tab:loss-ablations-updated}
\end{table}

\begin{table}[t]
    \centering
    \resizebox{1.\columnwidth}{!}{
    \setlength{\tabcolsep}{2pt} %
    \renewcommand{\arraystretch}{1.} %
    \begin{tabular}{l|c c|r r r} 
    \toprule
        \textbf{Method} & $A_S$ & $V_n$ & \textbf{Params (M)} & \textbf{GFLOPs} & \textbf{Inf. Time (ms)}\\
    \midrule
    AV-RIR \cite{AV-RIR}   & \cmark & \cmark & 390.66 & 270.43 & 794.06 \\
    M-CAPA (Ours)    & \cmark & \cmark   & \textbf{10.56} & \textbf{17.98} & \textbf{114.22}\\
    \midrule
    Image2Reverb \cite{image2reverb} & & \cmark& 57.6 & 276.91 & 198.44\\
    FAST-RIR\cite{Fast-RIR}++ & & \cmark& 132.68 & 57.84 & 121.76\\
    M-CAPA (Ours) & & \cmark& \textbf{5.84} &\textbf{ 11.24} & \textbf{76.61}\\
    \bottomrule
    \end{tabular}
    }
    \caption{Computational cost of the baselines and M-CAPA. Our approach is significantly faster and lighter. Lower is better for all metrics.
    \vspace{-0.5cm}}
    \label{tab:computational_metrics}
\end{table}

\subsection{Computational Cost} \label{sec:supp_compcost}
Our M-CAPA is a light-weight and efficient end-to-end model that can render RIRs conditioned on material profiles. Table \ref{tab:computational_metrics} compares the number of trainable parameters, GFLOPs, and inference time of M-CAPA to other SoTA approaches. Our model is significantly faster and lighter than the baselines.

\subsection{Performance Analysis on $D_{us}$ and $D_{uk}$}\label{sec:supp_perf} 

\begin{figure}[t]
    \centering
    \begin{subfigure}[t]{0.49\columnwidth}
        \centering
        \includegraphics[width=\linewidth]{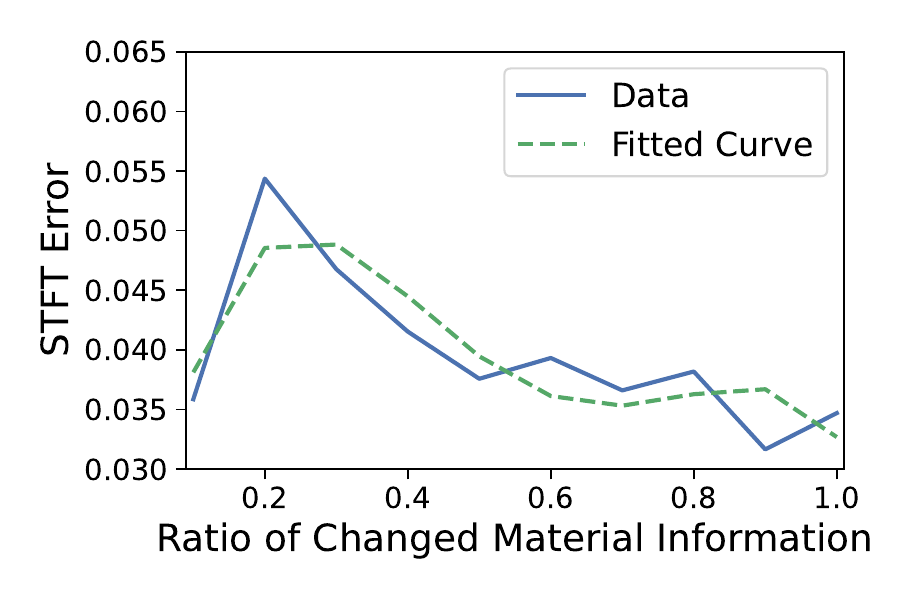} 
        \caption{}
        \label{fig:area_stft_dus}
    \end{subfigure}%
    \hfill
    \begin{subfigure}[t]{0.49\columnwidth}
        \centering
        \includegraphics[width=\linewidth]{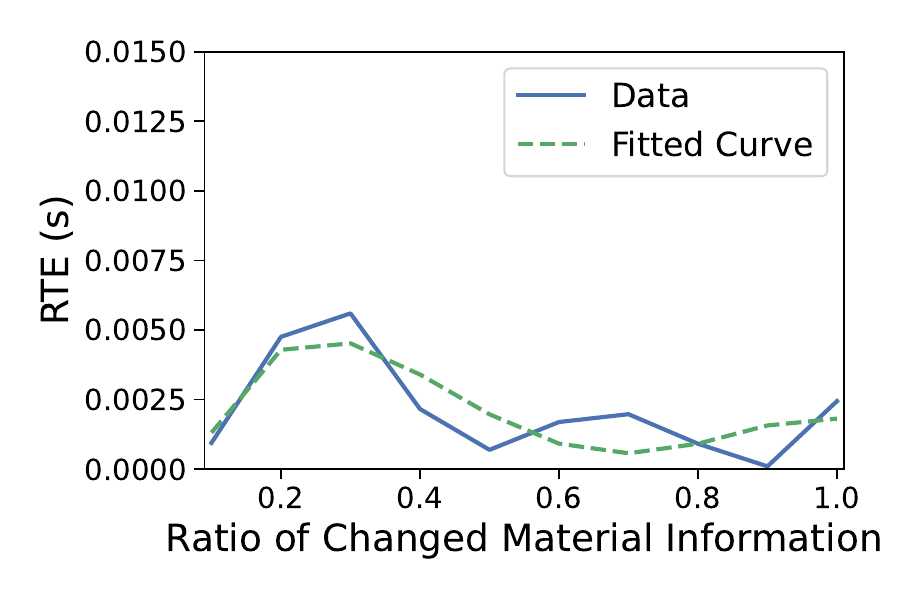} 
        \caption{}
        \label{fig:area_rte_dus}
    \end{subfigure}
    \\
    \begin{subfigure}[t]{0.49\columnwidth}
        \centering
        \includegraphics[width=\linewidth]{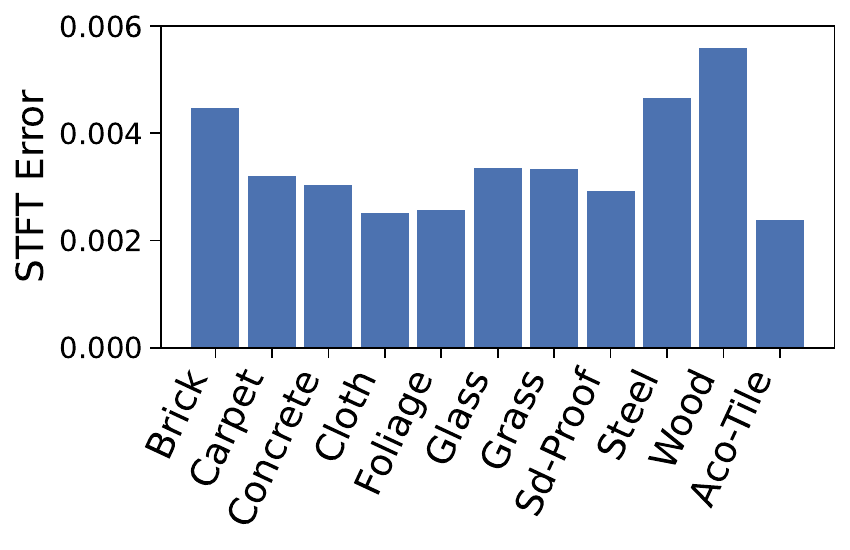} 
        \caption{}
        \label{fig:mat_stft_dus}
    \end{subfigure}%
    \hfill
    \begin{subfigure}[t]{0.49\columnwidth}
        \centering
        \includegraphics[width=\linewidth]{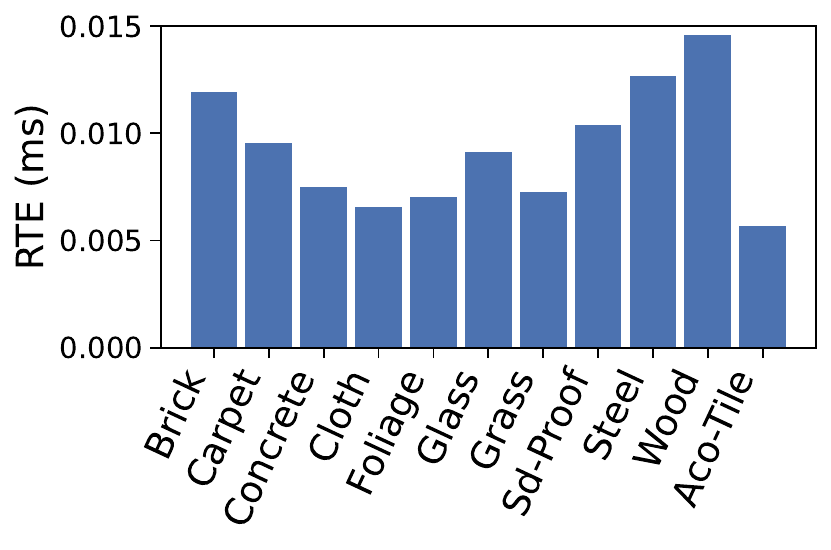} 
        \caption{}
        \label{fig:mat_rte_dus}
    \end{subfigure}
    \vspace{-0.2cm}
    \caption{
        Performance analysis of our model on $D_{us}$ with respect to the percentage of new material assignments in $\mathcal{M}_T$ (a and b) and across different material classes (c and d).
    }
    \label{fig:error-dus}
    \vspace{-0.5cm}
\end{figure}

\begin{figure}[t]
    \centering
    \begin{subfigure}[t]{0.49\columnwidth}
        \centering
        \includegraphics[width=\linewidth]{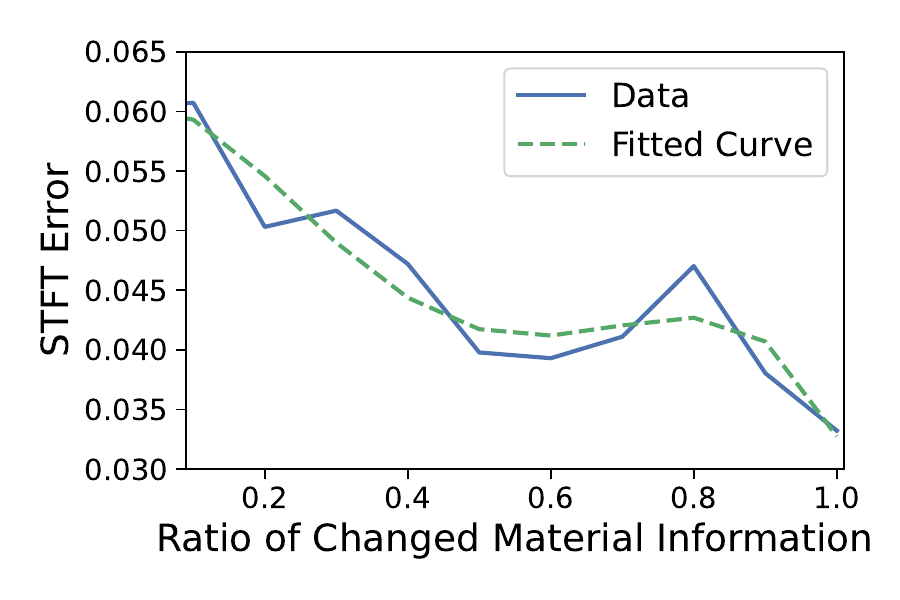} 
        \caption{}
        \label{fig:area_stft_duk}
    \end{subfigure}%
    \hfill
    \begin{subfigure}[t]{0.49\columnwidth}
        \centering
        \includegraphics[width=\linewidth]{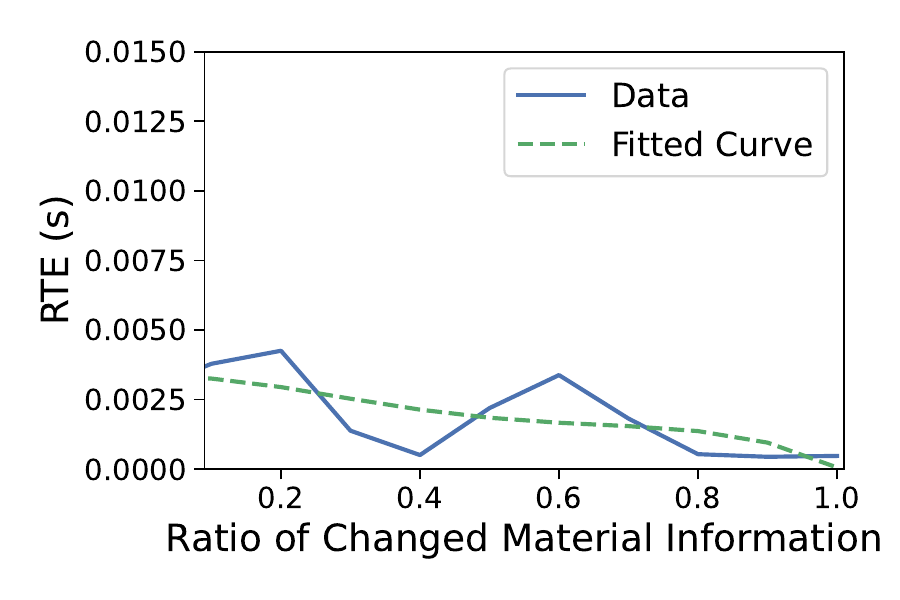} 
        \caption{}
        \label{fig:area_rte_duk}
    \end{subfigure}
    \\
    \begin{subfigure}[t]{0.49\columnwidth}
        \centering
        \includegraphics[width=\linewidth]{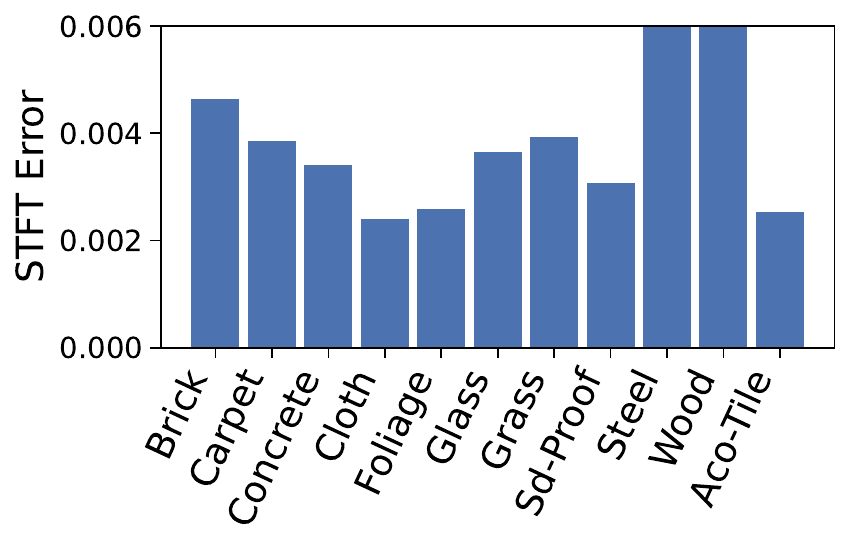} 
        \caption{}
        \label{fig:mat_stft_duk}
    \end{subfigure}%
    \hfill
    \begin{subfigure}[t]{0.49\columnwidth}
        \centering
        \includegraphics[width=\linewidth]{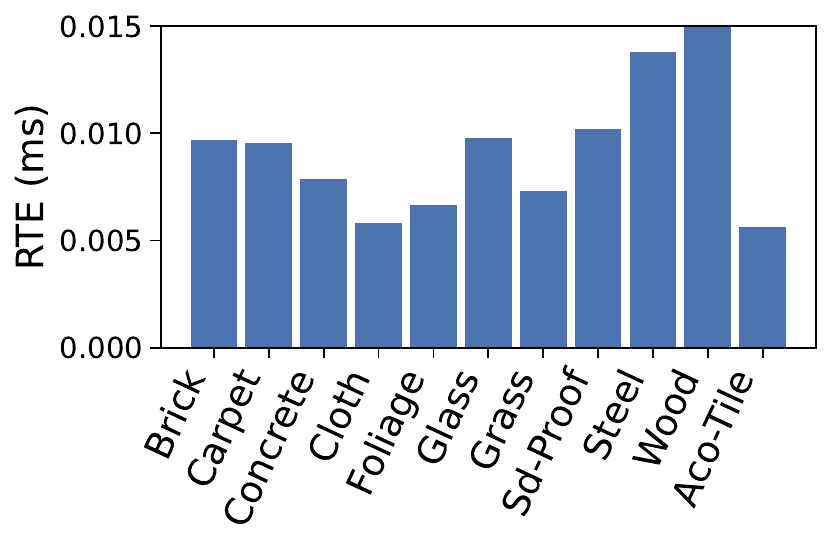} 
        \caption{}
        \label{fig:mat_rte_duk}
    \end{subfigure}
    \vspace{-0.2cm}
    \caption{
        Performance analysis of our model on $D_{uk}$ with respect to the percentage of new material assignments in $\mathcal{M}_T$ (a and b) and across different material classes (c and d).
    }
    \label{fig:error-duk}
    \vspace{-0.5cm}
\end{figure}

We analyze the performance of our model with respect to the changed material area in $\mathcal{M}_T$ and the different material classes, on the remaining test splits $D_{us}$ (Fig.~\ref{fig:error-dus}) and $D_{uk}$ (Fig.~\ref{fig:error-duk}). In both cases, we observe that our model generally benefits from material changes applied to larger areas within the scene. Larger areas provide more information to the model about how the target acoustic profile may change, compared to cases where only a small area undergoes new material assignments. 

Furthermore, consistent with our analysis of performance on $D_{uu}$, we find that certain material classes, such as \emph{Steel} and \emph{Wood}, are relatively more challenging for the model to accurately predict compared to others.

\subsection{Evaluation Results on Seen Environments}\label{sec:supp_eval_seen} %

We present the performance of our model in \textit{seen} environments in Table~\ref{tab:supp_eval_ss}. These environments are observed during training, and we evaluate performance under two setups: with seen material profiles ($D_{ss}$) and with unseen material profiles ($D_{su}$). The results for the split where both environments and materials match the training setup ($D_{ss}$) show that baselines, such as the Material Aware baseline, perform exceptionally well. This is expected, as both the evaluation and training samples originate from the same scene and material distributions, enabling these baselines to overfit effectively to the training data. However, this overfitting results in poor generalization to unseen material profiles ($D_{su}$), as shown in the left side of Table~\ref{tab:supp_eval_ss}, and limited generalization to unseen environments, as highlighted in the main experiments (Table~\ref{tab:main}). In contrast, our model, \emph{M-CAPA}, demonstrates robust generalization across unseen material profiles and unseen environments, as demonstrated by the results. %

\begin{table}[t]
    \centering
    \tiny
    \resizebox{\columnwidth}{!}{
    \setlength{\tabcolsep}{2pt} %
    \renewcommand{\arraystretch}{1.} %
    \begin{tabular}{l|c c|r r r r|r r r r} 
    \toprule
        & \multicolumn{10}{c}{\textbf{Seen Environments}}\\
     & \multicolumn{2}{c|}{\textbf{Observation}} & \multicolumn{4}{c|}{\textbf{Seen Materials} ($D_{ss}$)} & \multicolumn{4}{c|}{\textbf{Unseen Materials} ($D_{su}$)}\\
    \textbf{Method}  &\textbf{$A_s$} & \textbf{$V_n$} & \multicolumn{1}{c}{\textbf{L1}} & \multicolumn{1}{c}{\textbf{STFT}} & \multicolumn{1}{c}{\textbf{RTE}} & \multicolumn{1}{c|}{\textbf{CTE}} & \multicolumn{1}{c}{\textbf{L1}} & \multicolumn{1}{c}{\textbf{STFT}} & \multicolumn{1}{c}{\textbf{RTE}} & \multicolumn{1}{c}{\textbf{CTE}}  \\      %
    \midrule
    Direct Mapping    &    \cmark &         & 8.22 & 8.29 & 121.01 & 12.07 & 8.33 & 8.27 & 120.97 & 12.99   \\ 
      {M-CAPA (Ours)}   &    \cmark &         &   {\textbf{5.96}} &   {\textbf{4.63}} &   {\textbf{92.33}} &   {\textbf{7.73}} &   {\textbf{5.98}} &   {\textbf{4.62}} &   {\textbf{93.96}} &   {\textbf{8.72}}   \\ 
    \midrule
      {Image2Reverb\cite{image2reverb}}  &     &      \cmark        &   {14.35} &   {7.60} &   {253.02} &   {20.95} &   {14.12} &   {7.39} &   {237.69} &   {21.48}  \\ %
    
      {FAST-RIR++\cite{Fast-RIR,majumder2022fsrir}}   &     &      \cmark             &   {17.25} &   {32.45} &   {303.95} &   {22.95} &   {17.21} &   {33.51} &   {316.15} &   {21.91}  \\ %
    Material Agnostic     &     &      \cmark           & 8.18 & 8.11 & 119.23 & 11.47 & 8.23 & 8.24 & 117.03 & 12.33  \\
    Material Aware       &     &      \cmark            & {\bf3.47} & {\bf3.36} & \textbf{57.68} & {\bf5.09} & 7.27 & 7.02 & \textbf{83.91} & 9.79  \\ 
      {M-CAPA (Ours)}   &    &   \cmark       &   {5.98} &   {5.17} &   {90.16} &   {7.62} &   {\textbf{5.96}} &   {\textbf{5.05}} &   {91.59} &   {\textbf{8.64}}   \\ 
    \midrule
    AV-RIR~\cite{AV-RIR}  &  \cmark   &   \cmark            & 7.66 & 8.14 & \textbf{64.47} & 10.56 & 8.16 & 8.22 & \textbf{85.83} & 11.67  \\ 
    M-CAPA (Ours)          &  \cmark   &   \cmark         & \textbf{5.80} & \textbf{4.63} & 90.72 & \textbf{7.71} & {\bf5.81} & {\bf4.61} & 91.56 & {\bf8.70}\\ 
    \bottomrule
    \end{tabular}
    }
    \caption{
        Results on seen environments (used during training) when evaluated under two conditions: when coupled with seen material profiles ($D_{ss}$) which match exactly the training setup, and when coupled with unseen material profiles ($D_{su}$). Certain methods, such as the Material Aware, appear to overfit to the training samples in $D_{ss}$, leading to poor generalization performance on unseen cases like those in $D_{su}$. In contrast, our model, M-CAPA, demonstrates better generalization capabilities, achieving improved performance on $D_{su}$ while maintaining balanced results on $D_{ss}$.
        STFT and $L_1$ are scaled by $\times 10^{-2}$, RTE is in milliseconds (ms), and CTE in decibels (dB). Lower values indicate better performance for all metrics.
    }

    \label{tab:supp_eval_ss}
\end{table}

\subsection{Noise Experiments}\label{sec:supp_noise}

\begin{figure}[t]
    \centering
    \begin{subfigure}[t]{0.49\columnwidth}
        \centering
        \includegraphics[width=\linewidth]{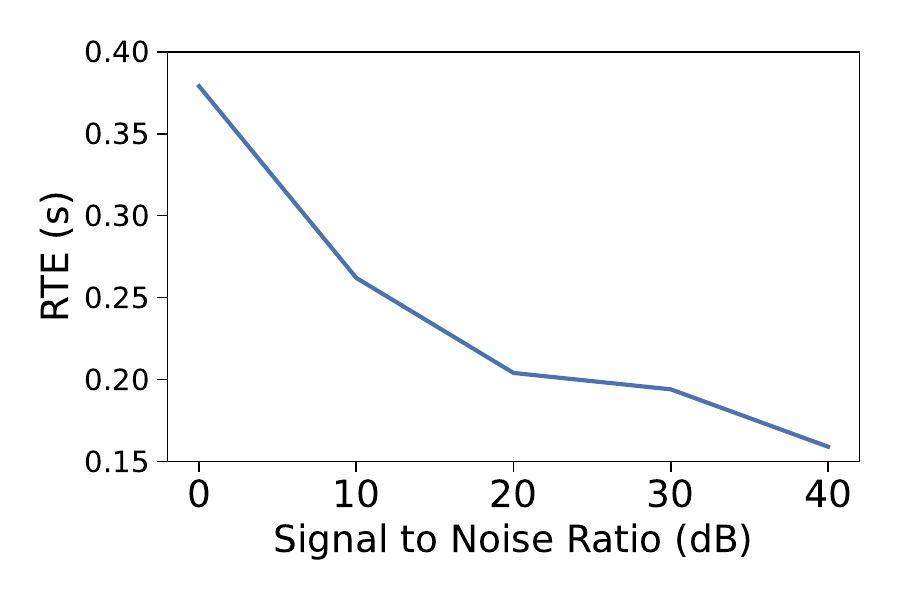} 
        \caption{}
        \label{fig:noise_stft}
    \end{subfigure}%
    \hfill
    \begin{subfigure}[t]{0.49\columnwidth}
        \centering
        \includegraphics[width=\linewidth]{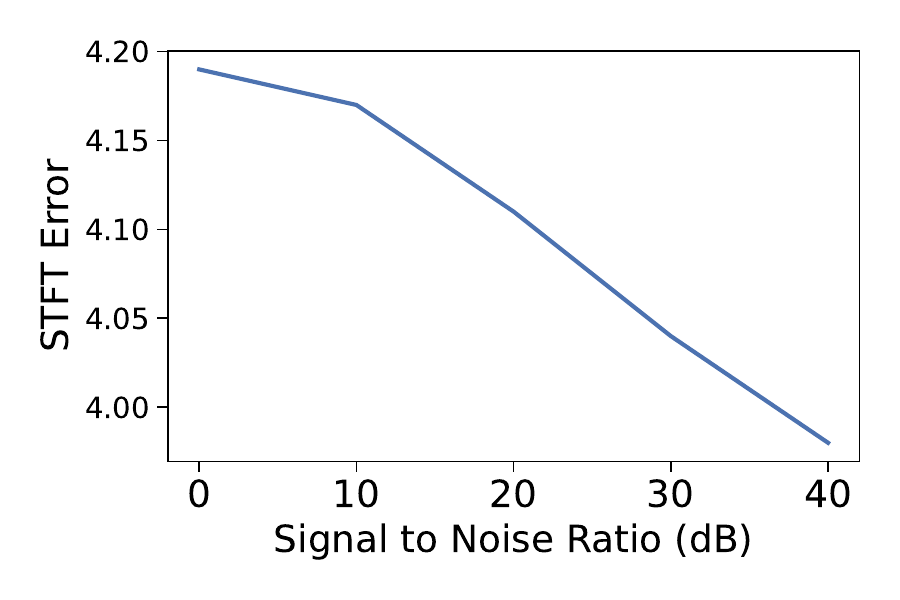} 
        \caption{}
        \label{fig:noise_rte}
    \end{subfigure}
    \vspace{-0.2cm}
    \caption{ 
        Robustness to noise. We introduce increasing levels of noise to the source RIR $A_S$ during inference, ranging from an SNR of 40 dB (clean $A_S$) to 0 dB (extremely noisy $A_S$), and evaluate performance on $D_{uu}$. For both metrics, lower values indicate better performance. %
    }
    \label{fig:noise}
    \vspace{-0.5cm}
\end{figure}

We evaluate the robustness of our model against noisy estimates of $A_S$. During inference, we introduce Gaussian noise to the source RIR with varying levels of strength, ranging from a signal-to-noise ratio (SNR) of 40 dB (relatively clean $A_S$) to 0 dB (extremely noisy $A_S$). In Fig.~\ref{fig:noise}, we illustrate the impact of noise on our model's performance for both the STFT error and the RTE metrics on the $D_{uu}$ split (a similar trend is observed on the other test splits). 

Our results show that the model's performance degrades gradually as the noise level increases. We believe that the robustness of our model to noise could be improved by incorporating data augmentation techniques with noisy inputs during training. We leave this as a direction for future work.

\begin{figure*}[th!]
    \centering
    \includegraphics[width=0.9\linewidth]{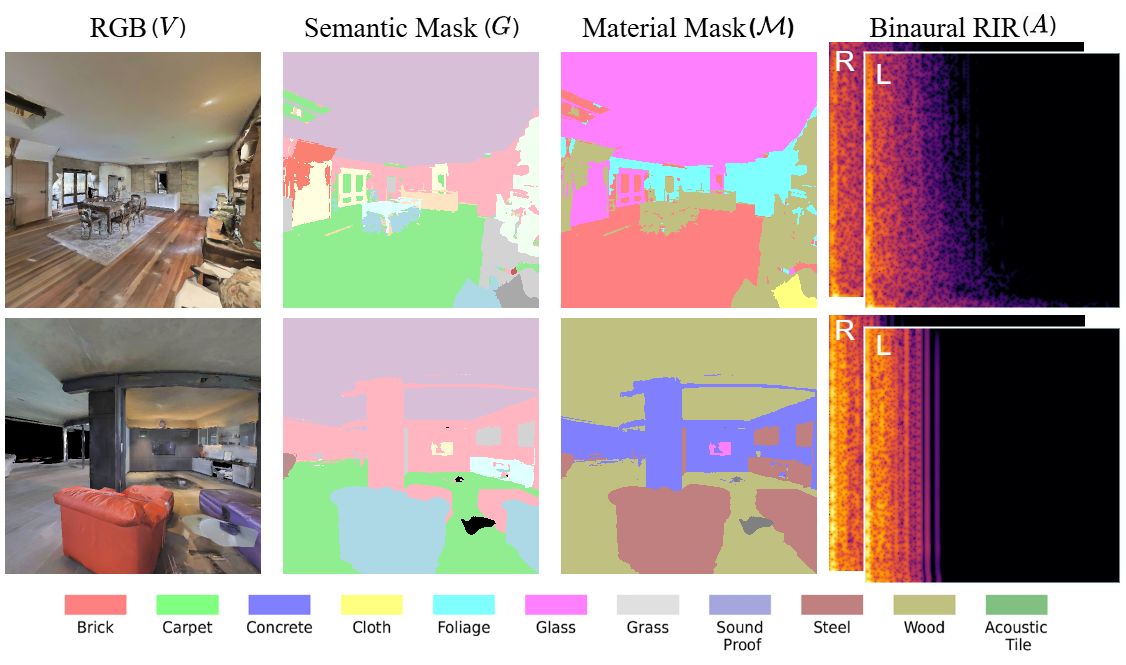}
    \caption{
        Examples from our Acoustic Wonderland Dataset. Each data point contains an RGB image, a semantic segmentation mask, a material segmentation mask, and the corresponding acoustic profile in the form of a two-channel RIR.  
    }
    \label{fig:dataset_examples}
\end{figure*}

\subsection{Acoustic Wonderland Dataset}\label{sec:supp_datasest}
We provide detailed information regarding the creation and characteristics of our dataset, including the location sampling methodology, material properties, material profiles, and their pairings.

\paragraph{Location Sampling}
The locations for sampling data points in our dataset are selected based on specific criteria to ensure that each point lies in an open space within the environment and provides meaningful visual and acoustic information. 
The sampling process involves randomly selecting locations within an indoor scene, subject to the condition that no two sampled locations are closer than a predefined distance threshold of \textit{0.1m}.
This prevents sampling overlapping locations and ensures a more uniform spatial coverage of the scene. At each selected location, we place a sensor suite consisting of a camera, a speaker, and binaural microphones with a random orientation. To enhance the diversity and realism of the dataset, care is taken to avoid situations where the camera is positioned too close to, or directly facing, large objects such as walls or doors.

\vspace{-0.3cm}
\paragraph{Material Classes}
Our dataset incorporates 12 material classes, including \textit{wood}, \textit{steel}, \textit{concrete}, \textit{grass}, \textit{foliage}, \textit{glass}, \textit{brick}, \textit{steel}, \textit{sound-proof}, \textit{carpet} and \textit{acoustic tiles}. We also include a \textit{default} material class which is SoundSpaces default material mapped onto any unlabeled object in the scene. 
Each material class is characterized in SoundSpaces by its acoustic coefficients, such as reflection, absorption, transmission, and damping properties across various frequency bands of sound waves. These coefficients are essential for accurately modeling the acoustic behavior of the materials within the simulated environment.

\vspace{-0.3cm}
\paragraph{Material Profiles}
Each profile defines a mapping between material classes and semantic object categories within a scene. The SoundSpaces simulator utilizes this mapping to assign materials to objects based on their semantic labels. For each material profile in our dataset, a random mapping is generated to disentangle the relationship between material and semantic classes.
For instance, one material profile may assign \textit{wall} and \textit{floor} to the material \emph{wood}, while another profile maps \textit{wall} to \emph{concrete} and \textit{floor} to \emph{carpet}. These mappings are applied to large objects and surfaces, such as furniture, doors, and walls, while smaller objects (e.g., sports equipment, utensils, televisions) retain their default materials. This distinction is made because smaller objects typically have negligible impact on the overall acoustic profile of the scene. In total, we generate \textit{2,673} unique material profiles for our dataset. See examples in Fig.~\ref{fig:dataset_examples}.

\paragraph{Pairings}
Following the observation sampling step described in the main paper (Sec.~\ref{sec:dataset}), we sample, for each location, a random pairing of two observations derived from different material profiles: $O_{n,S} = (V_n, G_n, \mathcal{M}_{n,S}, A_{n,S})$ and $O_{n,T} = (V_n, G_n, \mathcal{M}_{n,T}, A_{n,T})$. 
In this pairing, one observation serves as the source configuration, representing the original state of the scene $(V_n, G_n, A_{n,S})$, while the other represents the target state $(\mathcal{M}_{n,T}, A_{n,T})$, after applying a material change. The material change is denoted as $\mathnormal{diff}(\mathcal{M}_{n,T}, \mathcal{M}_{n,S})$. This setup simulates a scenario where a user alters the material configuration of the scene from $\mathcal{M}_{n,S}$ to $\mathcal{M}_{n,T}$, and the objective is to generate the corresponding target acoustic profile $A_{n,T}$.

\begin{table*}[t]
    \centering
    \tiny
    \resizebox{1.\textwidth}{!}{
    \setlength{\tabcolsep}{2pt} %
    \renewcommand{\arraystretch}{1.} %
    \begin{tabular}{l|c c|r r r r|r r r r|r r r r} 
    \toprule
    & \multicolumn{2}{c|}{\textbf{Observation}} & \multicolumn{4}{c|}{\textbf{Seen Materials}} & \multicolumn{4}{c|}{\textbf{Unseen Materials}} & \multicolumn{4}{c}{\textbf{Unseen Pairings}} \\
    \textbf{Method} & \textbf{$A_s$} & \textbf{$V_n$} &\multicolumn{1}{c}{\textbf{L1}} & \multicolumn{1}{c}{\textbf{STFT}} & \multicolumn{1}{c}{\textbf{RTE}} & \multicolumn{1}{c|}{\textbf{CTE}} & \multicolumn{1}{c}{\textbf{L1}} & \multicolumn{1}{c}{\textbf{STFT}} & \multicolumn{1}{c}{\textbf{RTE}} & \multicolumn{1}{c|}{\textbf{CTE}} & \multicolumn{1}{c}{\textbf{L1}} & \multicolumn{1}{c}{\textbf{STFT}} & \multicolumn{1}{c}{\textbf{RTE}} & \multicolumn{1}{c}{\textbf{CTE}} \\   
     \midrule
    Direct Mapping        & \cmark  &        & 9.63 & 10.29 & 132.7 & 14.65 & 9.59 & 10.31 & 134.4 & 15.04 & 9.97 & 10.89 & 133.9 & 14.03  \\ 
    M-CAPA (Ours)     & \cmark  &   & {\bf 6.75} & {\bf 5.38} & {\bf 98.28} & {\bf 9.05} & {\bf 6.76} & {\bf 5.42} & {\bf 102.2} & {\bf 9.41} & {\bf 7.25} & {\bf 6.12} & {\bf 100.2} & {\bf 9.91} \\ %
    \midrule
    Image2Reverb~\cite{image2reverb} & & \cmark & 18.38 & 9.51 & 234.1 &39.92 & 17.56  & 8.91 & 202.2 & 40.65 & 16.36 & 9.27 & 231.5 & 37.89 \\ 
    FAST-RIR++~\cite{Fast-RIR,majumder2022fsrir} & & \cmark &  18.97 & 34.88 & 311.4 & 20.78 & 18.67 & 37.29 & 324.8 & 20.30 & 19.71& 44.80 &312.0 & 20.67 \\ 
    Material Agnostic   & & \cmark            & 10.06 & 13.27 & 127.8 & 14.28 & 10.12 & 13.01 & 127.1 & 14.56 & 10.49 & 13.76 & 129.9 & 13.93 \\
    Material Aware        & &  \cmark          & 9.88 & 12.64 & 105.2 & 11.81 & 9.81 & 12.65 & 102.5 & 12.18 & 10.60 & 13.75 & 106.3 & 12.05 \\ 
    M-CAPA (Ours)  && \cmark& \textbf{7.16} & {\bf 7.23} & \textbf{96.90}& {\bf 9.30} &\textbf{7.13} & {\bf 7.23} &\textbf{98.28} & {\bf 9.65} &\textbf{7.70} & {\bf 8.24} & \textbf{101.3} & \textbf{10.03} \\ 
    \midrule
    AV-RIR~\cite{AV-RIR}   & \cmark & \cmark         & 9.62 & 10.30 & 108.3 & 12.78 & 9.57 & 10.32 & 106.7 & 12.78 & 10.06 & 10.97 & 107.4 & 12.35 \\ 
    M-CAPA (Ours)           & \cmark & \cmark        & {\bf 6.57} & {\bf 5.39} & \textbf{97.58} & \textbf{8.99} & {\bf 6.54} & {\bf 5.42} & {\bf 101.0} & {\bf 9.22} & {\bf 7.07} & {\bf 6.15} & \textbf{101.6} & {\bf 9.81}
    
    \\

    \bottomrule
    \end{tabular}
    }
    \caption{
        Results on unseen environments with ($A_S$,$A_T$) samples that have $L_2 \geq 75$ for our three test splits: $D_{us}$, $D_{uu}$ and $D_{uk}$. STFT and $L_1$ are scaled by $\times 10^{-2}$, RTE is in milliseconds (ms), and CTE in decibels (dB). Lower values indicate better performance for all metrics.
    }

    \label{tab:main-filtered}
\end{table*}

\begin{table}[t]
    \centering
    \resizebox{1.\columnwidth}{!}{
    \setlength{\tabcolsep}{2pt} %
    \renewcommand{\arraystretch}{1.} %
    \begin{tabular}{l|r r r r} 
    \toprule
        \textbf{Method} & \textbf{L1} & \textbf{STFT} & \textbf{RTE} & \textbf{CTE}\\
    \midrule
    M-CAPA (Ours)                &6.56 &5.42 & 101.0 & 9.25\\ 
    \midrule
     
    a) Ours w/o $\mathcal{M}_T$ & 6.91 & 5.64 & 117.1 & 10.02 \\ 
    b) Ours w/o $B_T$           & 7.20 & 6.98 & 116.3 & 12.52 \\ 
    c) Ours w/ Inferred $G_n$   & 6.93 & 5.56 & 107.3 & 9.96 \\ 
    d) Ours w/ Changed $\mathcal{M}_T$  & 6.78 & 5.59 & 108.1 & 9.91 \\ 
    
    \bottomrule
    \end{tabular}
    }
    \caption{Ablation of our model on the test split $D_{uu}$ with distance between ($A_S$, $A_T$) $\geq 75$. Lower is better for all metrics.
    }
    \vspace{-0.5cm}
    \label{tab:ablations-filtered}
\end{table}

\vspace{-0.3cm}
\paragraph{Perceptual Differences}
When collecting our dataset, we filtered out any samples in which less than 10\% of the input view contained changed material to ensure a noticeable difference between $A_S$ and $A_T$. However, does our data correspond to samples with noticeable perceptual differences observed by the users? To investigate this, we selected 45 samples uniformly from various $L_2$ differences between $A_S$ and $A_T$ in our test data, along with 15 controlled samples featuring identical RIR pairs where $A_S = A_T$. We then asked 8 users to listen to sounds convolved with both RIRs and determine whether they sounded the same or different.

\begin{figure}[t]
    \centering
    \begin{subfigure}[t]{0.49\columnwidth}
        \centering
        \includegraphics[width=\linewidth]{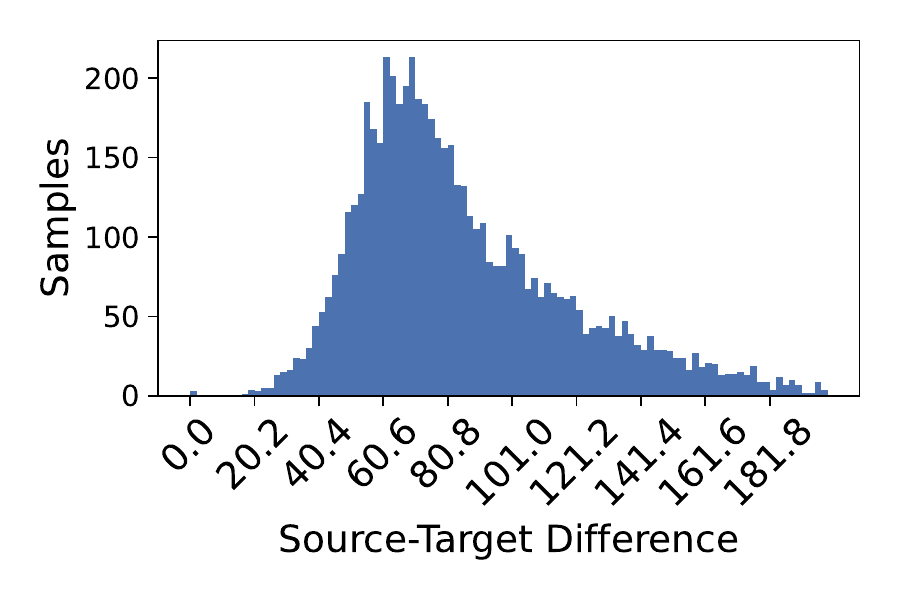} 
        \caption{}
        \label{fig:train-split-audio-diffs}
    \end{subfigure}%
    \hfill
    \begin{subfigure}[t]{0.49\columnwidth}
        \centering
        \includegraphics[width=\linewidth]{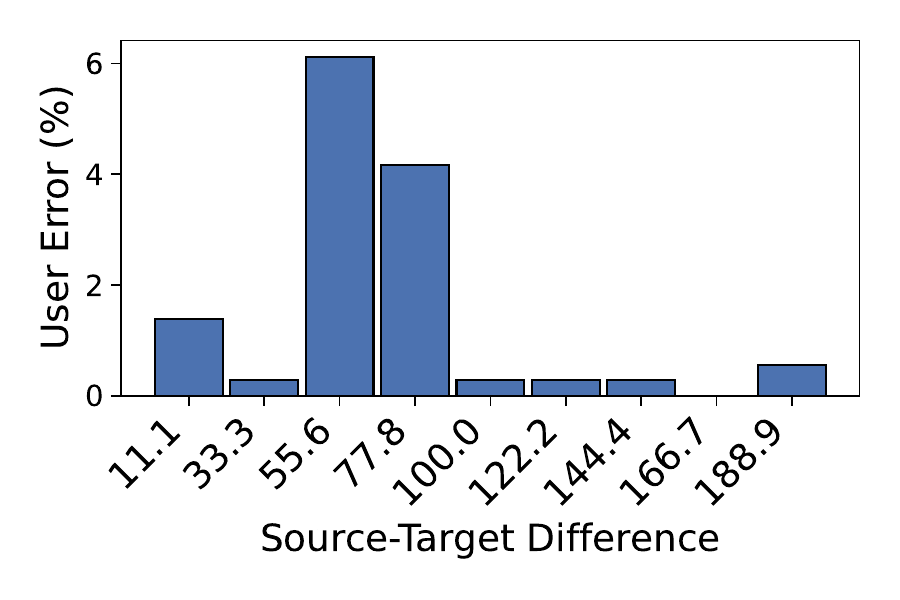} 
        \caption{}
        \label{fig:perceptual-user-study}
    \end{subfigure}
    \vspace{-0.2cm}
    \caption{ Analysis of perceptual differences in test data. Left, we show the distribution of differences between ($A_S$, $A_T$) in our unseen environments test splits. Right, we analyze the breakdown of errors accumulated by users during the perceptual difference user study. Overall, the error is low across all bins (below 6\%), and as the $L_2$ distance between $A_S$ and $A_T$ increases, perceptual differences become more apparent and user error decreases.
    }
    \label{fig:perceptual-diff-analysis}
    \vspace{-0.5cm}
\end{figure}

Our results show that the users achieved 87.9\% accuracy, indicating a strong perceptual distinction in our dataset. We show error distribution for the user study in Figure \ref{fig:perceptual-user-study}. Most errors occurred when the $L_2$ difference was in the lower range (11.1 to 77.8), suggesting that smaller variations in L2 distance are less perceptually salient. However, in general the error is low, below 6\%, across all $L_2$ bins.

In Table~\ref{tab:main-filtered} and Table\ref{tab:ablations-filtered}, we present the performance of different models on our test data, focusing only on samples with high perceptual differences ($L_2 \geq 75$). The results show that our model maintains its advantage over state-of-the-art and baselines in this setting as well.

\subsection{Model Architecture Details}\label{sec:supp_model}
The encoders in our model are based on a convolutional neural architecture inspired by the UNet~\cite{unet}. Each encoder ($f^V$, $f^G$, $f^A$, or $f^M$) comprises four downsampling layers. Each layer includes a convolutional block followed by a downsampling module.

The convolutional block consists of two consecutive Conv2D layers, each with a kernel size of 3, a batch normalization layer, and a LeakyReLU activation~\cite{xu2015empirical}. To enhance generalization, a dropout layer ~\cite{srivastava2014dropout} with a rate of 0.2 is included in each layer.

The downsampling module within each encoder layer consists of a MaxPooling layer with a kernel size of 2 and a stride of 2. This reduces the spatial resolution by a factor of 2 at each layer. The four layers of the encoder use 32, 64, 128, and 512 kernels, respectively. 

The fusion layer, $\mathcal{F}$, combines the multimodal scene embedding $e_m$ and the material embedding $e_t$. This fusion is performed using a single Conv2D layer with a kernel size of 3 and a stride of 1, which effectively integrates information from both embeddings into a unified representation.

The decoder, $f^T$, follows an architecture similar to the encoders but in a mirrored configuration. It consists of four upsampling blocks. Each upsampling block contains a single Transpose Conv2D layer, followed by two Conv2D layers, a batch normalization layer, and a LeakyReLU activation function. Skip connections are incorporated from the corresponding layers of the $f^A$ encoder, allowing the decoder to leverage features from earlier stages of the encoding process. The final output of the decoder is a two-channel binaural magnitude spectrogram of the target acoustic response.

\subsection{Evaluation Setup}\label{sec:supp_eval_setup}
In this section, we provide additional details about the baselines and evaluation metrics used in our experiments.

\paragraph{Baselines}
\begin{itemize}[leftmargin=*,align=left]  
    \item \textbf{Direct Mapping}: This baseline directly uses $A_S$ as the prediction for $A_T$, effectively ignoring the target material information. In other words, it assumes that the original acoustic response is sufficient to predict the target response. This baseline serves as a reference for quantifying the impact of material configuration on the target acoustics, as $A_S$ already captures the scene shape, object distribution, and original material configuration.
    
    \item \textbf{Material Agnostic Matcher}: In this baseline, we compute the cosine similarity between the visual embedding $e_v$ of the input and the embeddings of visual observations $V_n$ in the training set. The most similar data point is selected, and a random RIR associated with that location $l_n$ is returned as the prediction. This approach represents methods that estimate RIRs based on visual characteristics of the scene alone, without incorporating material information.
    
    \item \textbf{Material Aware Matcher}: Similar to the Material Agnostic Matcher, this baseline identifies the most visually similar scene location $l_n$ from the training data. However, in addition to visual similarity, it takes material information into account. From the set of RIRs associated with different material profiles at the selected location, we compute the L1 distance between the material distribution associated with each RIR and the target material distribution $\mathcal{M}_T$. The RIR with the most similar material distribution to $\mathcal{M}_T$ is selected. This baseline highlights the importance of accounting for material configuration and the similarity between material settings during training and testing.
    
    \item \textbf{Image2Reverb}~\cite{image2reverb}: We follow the official implementation provided by the authors to train this model on our dataset. With the same pre-trained depth and visual encoders from the original implementation, we train the GAN-based network to predict RIRs using the Acoustic Wonderland dataset. %
    
    \item \textbf{AV-RIR}~\cite{AV-RIR}: The AV-RIR model initially infers the RIR from reverberant speech and then estimates the late components of the RIR using a retrieved sample from a material-aware training database. To adapt this baseline to our case and improve its performance, we make the following changes:
    (1) Instead of inferring the source RIR from reverberant speech, we provide $A_S$ directly as input, as it offers a more accurate representation;
    (2) Similar to the Material Aware Matcher baseline, we retrieve the RIR of the closest training sample based on both visual and material-based similarity to the input sample.
      {(3) While the original implementation uses  a $\mathbf{360^\circ}$} panoramic RGB images to predict target RIRs, we choose to retrieve the closest sample in the training set using $\mathbf{90^\circ}$ Field of View (FoV) for fair comparison with M-CAPA which also uses $\mathbf{90^\circ}$ FoV. When comparing the impact of FoV on the performance of the AV-RIR baseline, we note that an increased FoV yields only marginal improvement. For example, in test split $D_{uu}$, L1 error drops from 7.59 to 7.49, STFT error reduces from 7.17 to 7.12, RTE improves from 99.10ms to 98.56ms and CTE drops from 11.35 to 11.22. This suggests that $A_S$ already carries significant cues about the entire room, without needing $\mathbf{360^\circ}$ FoV as visual input.   Following the AV-RIR approach, we retain the first 2000 samples of $A_S$ and replace the remaining samples with the reverberant components of the retrieved RIR.

      {\item \textbf{FAST-RIR++}: \cite{Fast-RIR} is a GAN-based approach to RIR synthesis for rectangular rooms, using properties of the acoustic environment such as room size, speaker/listener positions and reverberation time of the target RIR. We modify this approach following \cite{majumder2022fsrir} by making the following changes: 
    (1) Instead of providing the room size, we provide ground truth depth images, making this a vision-based variation of the original implementation.
    (2) In addition to RT60 provided by the original implementation, we also provide the direction-to-reverberant ratio (DRR) as an acoustic parameter of the room. We obtain acoustic parameters from the source RIR. We train FAST-RIR++ on our training dataset until convergence and evaluate on test splits. }
     
\end{itemize}

These baselines and existing methods address various aspects of evaluation and represent key directions in the RIR prediction literature. The \emph{Direct Mapping} baseline evaluates methods that focus solely on capturing the geometric and structural properties of the scene, without accounting for material changes. In contrast, the \emph{Material Agnostic} and \emph{Material Aware} baselines represent robust nearest-neighbor approaches. These baselines rely on the similarity between test and training scenes, either based purely on visual information or incorporating material representations. This comparison enables us to evaluate whether a method merely memorizes training data and whether the inclusion of material information leads to improved predictive performance.

Furthermore, \emph{Image2Reverb}, \emph{FAST-RIR++}, and \emph{AV-RIR} represent state-of-the-art (SoTA) approaches for RIR prediction. \emph{Image2Reverb} relies exclusively on visual inputs to predict the RIR of a scene. Interestingly, our findings reveal that \emph{Image2Reverb} demonstrates low performance in evaluations  {, even after retraining on our dataset}, being outperformed by  some of the baselines  in RTE and CTE.   {This observation shows that reliance on just RGB observations is not sufficient to render accurate RIRs that model material changes in the environment.   {\emph{AV-RIR} integrates material information within a more advanced prediction framework, estimating RIRs from reverberant speech, and finally conditioning late components of the estimated RIR using scene-based retrieval. AV-RIR focuses on limited material-object mapping, while our approach assumes all semantic objects in the scene are mapped to materials and contribute to the final RIR prediction. \emph{FAST-RIR++} provides an acoustically guided approach to RIR prediction, using target acoustic parameters to guide RIR generation. This baseline examines the impact of explicit acoustic parameters for the prediction of accurate RIRs. }}

\paragraph{Metrics}
We used the following metrics to evaluate performance: 
\begin{itemize}[leftmargin=*,align=left]  
    \item \textbf{L1 Error}: The L1 norm between the generated $\hat{A}_T$ and ground truth $A_T$ audio's magnitude spectrograms.
    \item \textbf{STFT Error}: The mean squared error (MSE) between the magnitude spectrograms of the generated and ground truth audio's magnitude spectrograms.
    \item \textbf{RTE}: This metric (Reverberation Time Error) quantifies the difference in time taken for the energy of the predicted signal $\hat{A}_T$ and the ground truth signal $A_T$ to decay by 60~dB. This is a standard metric used in prior works (e.g.,~\cite{Adverb, AV-RIR, majumder2022fsrir}). Following the approach in~\cite{majumder2022fsrir}, we use the Schroeder Integration Method~\cite{lentz2007virtual} to estimate the decay time. For binaural RIRs, we compute the reverberation time for both channels and report the average absolute difference between $\hat{A}_T$ and $A_T$.
    \item \textbf{CTE~\cite{vigran2014building}}: This metric 
    calculates the difference in the ratio of direct energy (the first 50~ms of the signal) to late energy for both signals, providing insight into how accurately a model captures the acoustic characteristics of the environment.
\end{itemize}

\paragraph{Signal Reconstruction}
For both RTE and CTE, a waveform representation of $\hat{A}_T$ is required. Reconstructing the target signal accurately necessitates the inclusion of phase information. To address this, we leverage the phase information from the source impulse response ($A_S$). By carrying over the phase from $A_S$, the predicted magnitude can be reconstructed into a waveform that can be directly compared to the target waveform, ensuring a meaningful evaluation of the reconstruction accuracy.

\end{document}